\documentclass{bmvc2k}
\usepackage{algorithm}
\usepackage{algorithmic}
\usepackage{amsfonts}       
\usepackage{amsmath}
\usepackage{bbm}
\usepackage{comment}
\usepackage{multirow}
\usepackage{booktabs,colortbl,tabularx}

\newcommand{\red}[1]{\textcolor{red}{#1}}

\def\thefootnote{*}\footnotetext{These authors contributed equally to this work. $^\dagger$Corresponding author.}

\def\corfootnote{$^\dagger$}

\title{Point3D: tracking actions as moving points with 3D CNNs}

\addauthor{Shentong Mo\thefootnote{}}{shentonm@andrew.cmu.edu}{1}
\addauthor{Jingfei Xia\thefootnote{}}{jingfeix@andrew.cmu.edu}{1}
\addauthor{Xiaoqing Tan}{xit31@pitt.edu}{1,2}
\addauthor{Bhiksha Raj\corfootnote{}}{bhiksha@cs.cmu.edu}{1}

\addinstitution{
 Carnegie Mellon University\\
 Pittsburgh, PA 15213, United States
}
\addinstitution{
 University of Pittsburgh\\
 Pittsburgh, PA 15260, United States
}

\runninghead{Shentong Mo, Jingfei Xia, Xiaoqing Tan, and Bhiksha Raj}{Point3D}

\begin{document}

\maketitle

\begin{abstract}
    Spatio-temporal action recognition has been a challenging task that involves detecting where and when actions occur.
    Current state-of-the-art action detectors are mostly anchor-based, requiring sensitive anchor designs and huge computations due to calculating large numbers of anchor boxes.
    Motivated by nascent anchor-free approaches, we propose \textbf{Point3D}, a flexible and computationally efficient network with high precision for spatio-temporal action recognition.
   Our Point3D consists of a \textit{Point Head} for action localization and a \textit{3D Head} for action classification.
   Firstly, Point Head is used to track center points and knot key points of humans to localize the bounding box of an action.
   These location features are then piped into a time-wise attention to learn long-range dependencies across frames.
   The 3D Head is later deployed for the final action classification.
    Our Point3D achieves state-of-the-art performance on the JHMDB, UCF101-24, and AVA benchmarks in terms of frame-mAP and video-mAP.
    Comprehensive ablation studies also demonstrate the effectiveness of each module proposed in our Point3D.
    
\end{abstract}
\vspace{-1.5em}
\section{Introduction}


Spatio-temporal action recognition has attracted much attention in the field of computer vision. It aims to locate and detect action instances of interest in a video in both space and time. The applications can range from human-computer interaction~\cite{rautaray2015vision, mitra2007gesture},  video surveillance~\cite{oh2011large, leo2004human}, to social activity recognition~\cite{coppola2019social, coppola2016social}.
Videos, unlike images, are time series of images that consist of both spatial components and temporal components. Hence, spatio-temporal action recognition requires both spatial information from key frames as well as temporal information from their previous frames~\cite{kopuklu2019yowo}. 

\begin{figure}[!htb]
\setlength{\abovecaptionskip}{-1.5em}
\setlength{\belowcaptionskip}{-0.5em}
		\centering
		\includegraphics[width=0.6\linewidth]{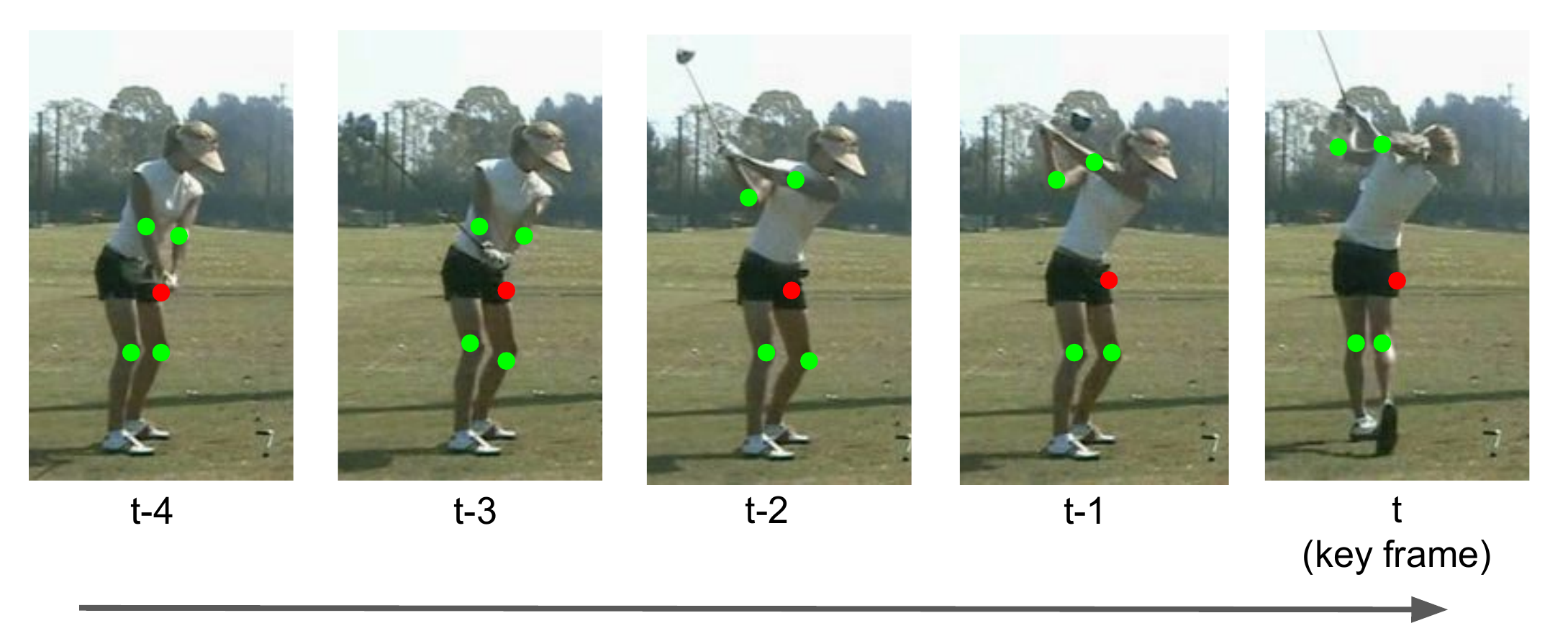}
    	\caption{\textbf{Motivation illustration}. Our proposed anchor-free network Point3D achieves high action detection precision by tracking the center point (red dot) and key points (green dots) with 3D-CNNs. The example figures are from the JHMDB dataset.}
	\label{fig: title_img}
\end{figure}



Early attempts for spatio-temporal action recognition typically apply an object detector at each frame of a video clip independently and then link those frame-wise detection results to generate action tubes \cite{gkioxari2015finding, peng2016multi, weinzaepfel2015learning, singh2017online}. 
However, those methods fail to consider temporal features across frames and are cumbersome to deploy in a real-time setting. 
Later, approaches~\cite{saha2017amtnet, kalogeiton2017action, hou2017tube, yang2019step, zhao2019dance, song2019tacnet, kopuklu2019yowo} aim to capture temporal information in videos by conducting clip-level action recognition. 
These tubelet detection methods, however, are anchor-based detectors that heavily rely on a large number of pre-defined anchor boxes as candidates for calculating bounding boxes. They can be highly sensitive and computationally inefficient as they exhaust all potential actor locations in an image and perform classification for each location~\cite{li2020actions}.

Recently, anchor-free approaches have gained increasing popularity in the computer vision community for their relatively simple network structures and computational efficiency \cite{li2020actions, law2018cornernet, duan2019centernet, tian2019fcos, zhou2019objects, zhou2020tracking}. By eliminating the anchor boxes, anchor-free methods avoids the intense computation as well as hyper-parameters tuning related to anchor boxes, resulting in a more robust performance \cite{li2020actions}. However, there has been little work on spatio-temporal action recognition under this framework.


Motivated by the nascent anchor-free framework, we propose a flexible and computationally efficient end-to-end detector called Point3D for the task of spatio-temporal action recognition. Our motivation is illustrated in Figure \ref{fig: title_img}.
Specifically, our Point3D performs the task of spatio-temporal action recognition by solving the localization and classification problems separately. Point3D consists of a \textit{Point Head} for action localization and a \textit{3D Head} for action classification. Specifically, the Point Head is used to track center points and knot key points of humans to localize the bounding box of an action. These location features are then piped into a time-wise attention mechanism to learn the dependencies between frames in a clip. The 3D Head is later deployed for the final action classification.
We conduct extensive experiments on the JHMDB and UCF101-24 datasets as well as ablation studies to demonstrate the effectiveness of each part of our network. 


The major contributions of this paper are summarized as follows: \textbf{1.} We present Point3D, an anchor-free end-to-end architecture with high precision for spatio-temporal action recognition. \textbf{2.} Our Point3D is the first of its kind to incorporate center points and knot key points together to localize the bounding box of an action without any anchors. \textbf{3.} We apply a time-wise attention mechanism to learn the temporal dependencies of each frame in an input clip to boost the performance. \textbf{4.} Our Point3D achieves competitive performance on the JHMDB, UCF101-24 and AVA datasets and extensive ablation studies show each part of our Point3D. 

\vspace{-1.5em}
\section{Related Works}
\vspace{-0.5em}

\subsection{Anchor-based Object Detection}
\vspace{-0.5em}
R-CNN \cite{girshick2014rich}, Fast R-CNN, \cite{girshick2015fast} and Faster R-CNN \cite{ren2015faster} are three of the pioneering works to introduce CNNs into the field of object detection. These three algorithms all fall into the category of two-stage object detectors that first 
generate regions of interest and then pipe those region proposals for object classification and bounding-box regression. 
The need for a two-stage network was eliminated by the YOLO architecture \cite{redmon2016you} where the task of object detection is regarded as a simple regression problem by taking an input image and learning its class probabilities and bounding box coordinates. 
Later improvements~\cite{redmon2016yolo9000, redmon2018yolov3, bochkovskiy2020yolov4} in YOLO involve optimizing small object localization, anchor boxes, and network structure. 


\vspace{-1em}

\subsection{Anchor-free Object Detection}

\vspace{-0.5em}

Recent advances in anchor-free object detection show surprisingly competitive performance without the need of computing expensive anchors \cite{law2018cornernet, zhou2019objects, zhou2020tracking, RepPointsv1}. 
CornerNet \cite{law2018cornernet} detects an object bounding box as a pair of key points using a single convolution neural network. 
CenterNet \cite{zhou2019objects} detects an object as a point in an image with a focus on modeling the center of a box as an object and uses this predicted center to get the bounding box coordinates. RepPoints \cite{RepPointsv1} instead uses multiple points to automatically learn the spatial extent of an object. CenterTrack \cite{zhou2020tracking} tracks objects from frame to frame by using the displacement of a point between adjacent frames. 




\vspace{-1em}

\subsection{Spatio-temporal Action Recognition}

\vspace{-0.5em}

Most spatio-temporal action detectors build their work in videos based on object detectors~\cite{wang2016actionness, hou2017tube, kalogeiton2017action, gu2018ava, sun2018actor, girdhar2019video, li2020actions, wu2019long}. Traditional anchor-based approaches conduct frame-level action detection by combining independent frame-level detection with linking algorithms to generate final tubes \cite{gkioxari2015finding, wang2016actionness, peng2016multi, singh2017online, weinzaepfel2015learning}. More recent works place more focus on incorporating temporal information to better the recognition of actions in videos \cite{saha2017amtnet, kalogeiton2017action, hou2017tube, yang2019step, zhao2019dance, song2019tacnet, kopuklu2019yowo,mo2020improving}.
Typically, 3D CNN~\cite{duarte2018videocapsulenet} and LSTM~\cite{he2017generic} are used to extract temporal features. AIA~\cite{tang2020asynchronous} also applies memory mechanism to model long-term interaction. 
However, many of these works apply a two-stage architecture, where proposals are first generated by RPN and then the tasks of action classification and localization are performed accordingly~\cite{peng2016multi, hou2017tube, wu2020contextaware}. 
Because the proposal network and the classification network are trained separately, the results may not be optimal. 
Plus, as the RPN only focuses on spatial information of actors in each frame, relationships between frames are largely neglected, failing to incorporate important temporal information for action classification and localization.

More recently, an end-to-end detector called YOWO~\cite{kopuklu2019yowo} has been proposed which uses a 2D-CNN branch and a 3D-CNN~\cite{kay2017kinetics} branch to extract spatial and temporal features respectively on video data. 
YOWO is flexible in that its 2D-CNN and 3D-CNN branches can be replaced by any CNN architectures of interest \cite{kopuklu2019yowo}. 
Inspired by anchor-free object detectors, MOC Detector~\cite{li2020actions} incorporates movement information along with center detection and spatial extent detection by treating an action instance as a trajectory of moving points. However, limited works have been proposed in the field of spatio-temporal action localization with anchor-free architectures. 
In this work, we propose an anchor-free end-to-end network, termed as Point3D, motivated jointly by the conceptually simple CenterNet~\cite{zhou2019objects}, the recent excellent performance of 3D-CNNs~\cite{kay2017kinetics}, as well as the flexible two-branch YOWO~\cite{kopuklu2019yowo}. Besides, our Point3D can be trained in an end-to-end manner, compactly incorporating both spatial and temporal information of a video clip. 
\setlength{\belowdisplayskip}{0.7pt} \setlength{\belowdisplayshortskip}{0.7pt}
\setlength{\abovedisplayskip}{0.7pt} \setlength{\abovedisplayshortskip}{0.7pt}

\vspace{-1.5em}
\section{Method}
\vspace{-0.5em}

\subsection{Overview}

\vspace{-0.5em}
In this work, we aim to solve the problems of localization of action bounding boxes in a video clip together with classification of the action categories. 
To address this problem, we propose an anchor-free architecture called Point3D, that operates by tracking actions as moving points and achieves better performance with the help of 3D-CNNs. 

\begin{figure}[t]
\setlength{\abovecaptionskip}{-1.5em}
\setlength{\belowcaptionskip}{-1.5em}
		\centering
		\includegraphics[width=0.65\textwidth]{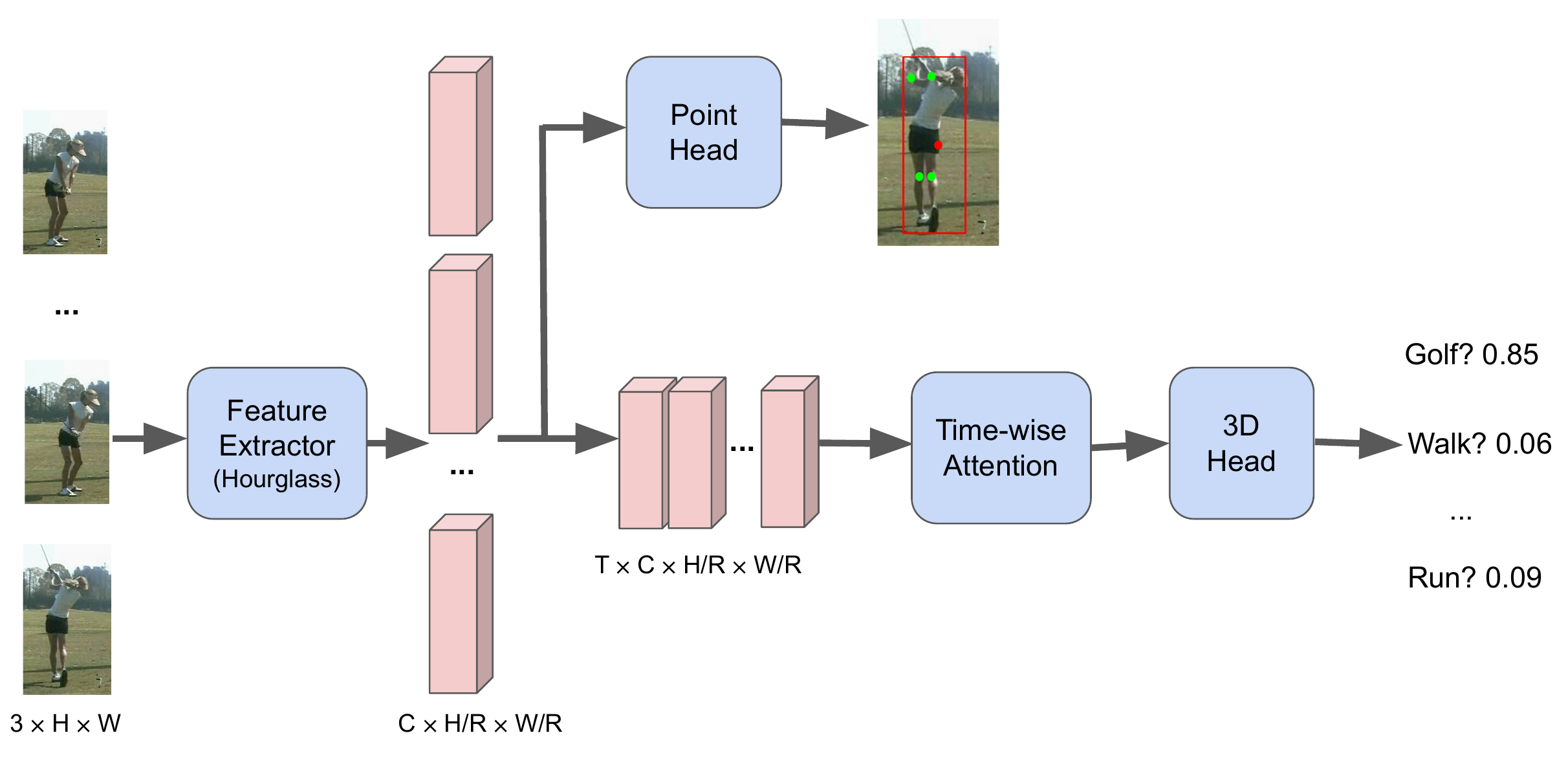}
    	\caption{\textbf{Point3D network overview.} An input clip is fed into Feature Extractor to obtain features of each frame. These features will then be used to solve the task of action location and classification separately. Point Head is applied to these features on each frame to track center points and knot key points of humans for action localization, where those features are piped into Time-wise Attention and a 3D Head for action classification.}
	\label{fig: overall_model}
\end{figure}

As illustrated in Figure~\ref{fig: overall_model}, Point3D consists of four components: 
(1) \textit{Feature Extractor}: an input clip is fed into Feature Extractor to obtain spatial features of each frame. 
(2) \textit{Point Head}: Point Head is applied for the localization task to track center points and knot key points of actors on each frame in a video clip.
(3) \textit{Time-wise Attention}: 
Time-wise Attention is designed to aggregate the correlation and continuity across frames for incorporating more context information.
(4) \textit{3D Head}: 
3D Head with a 3D-CNN backbone is 
applied to improve the performance of action classification.
Our Point3D is trained in an end-to-end manner, such that these components collaboratively work together to generate a more stable and accurate recognition in a video. 
We describe the technical details in the following sections.



\vspace{-1em}
\subsection{Feature Extractor}

\vspace{-0.5em}

Assume an input frame as $I \in \mathbb{R}^{3\times H \times W}$, where $H$ and $W$ are the height and width of the frame. 
There are $N$ actors in each frame and there are $T$ frames in a clip.
We implement a widely-used network in pose estimation, the Hourglass architecture~\cite{alejandro2016hourglass} as our backbone for feature extraction. 
Following~\cite{zhou2019objects}, the input frames are first downsampled by four times, and then they are passed through a two-stacked Hourglass structure to extract features for future localization and classification tasks. 
We denote the generated features of each frame to be of $\mathbf{f} \in \mathbb{R}^{C\times \frac{H}{R}\times \frac{W}{R}}$ where $C$ denotes number of channels and is set as 128, and $R$ denotes stride factor and is set as 4 following~\cite{li2020actions, zhou2019objects}. 

\vspace{-1em}
\subsection{Point Head}
\begin{figure}[!htb]
    \setlength{\abovecaptionskip}{-1.5em}
\setlength{\belowcaptionskip}{-1em}
		\centerline{\includegraphics[width=0.6\textwidth]{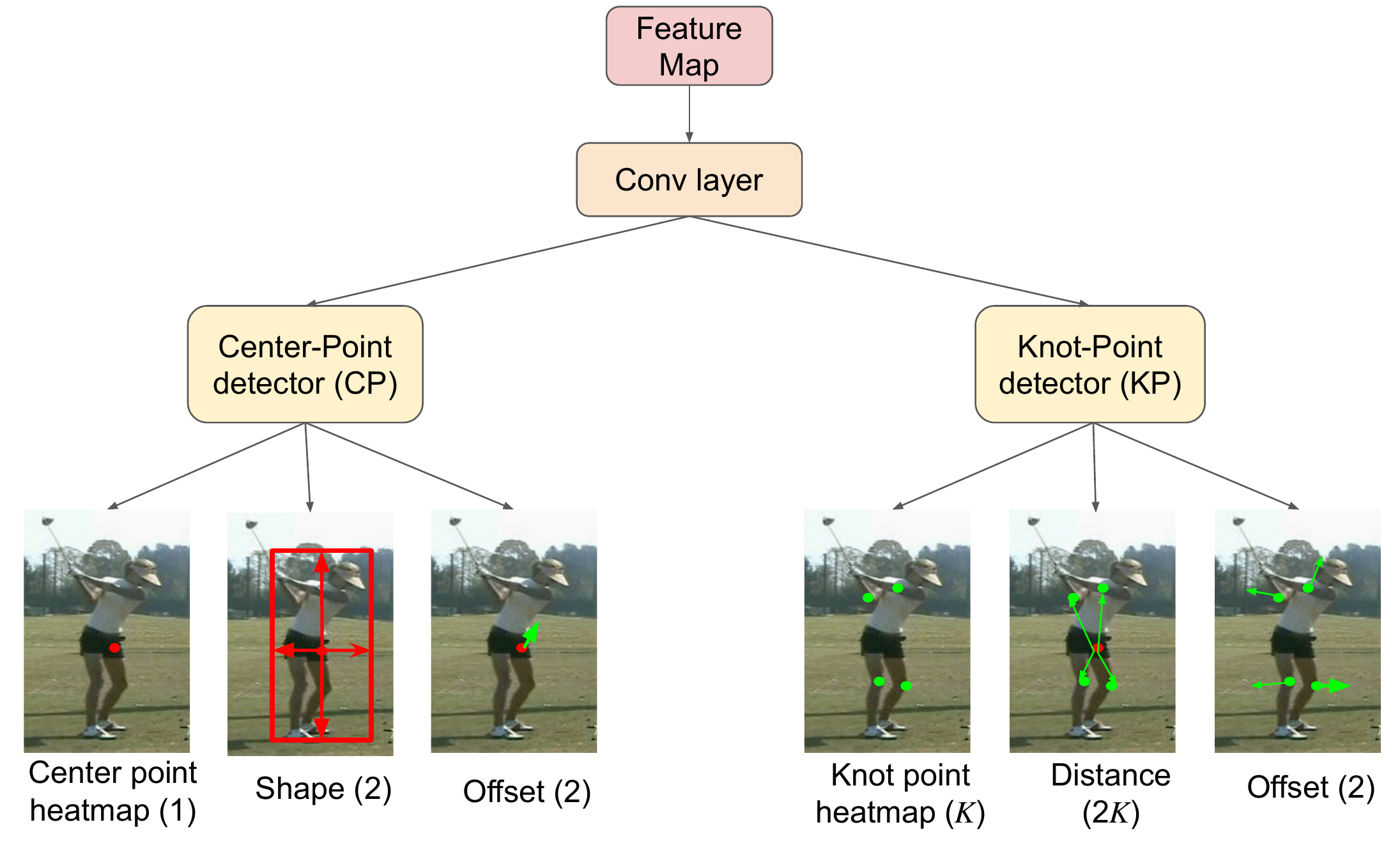}}
    	\caption{\textbf{Point Head structure} including CP detector (\textbf{Left}) and KP detector (\textbf{Right}). 
    	}
	\label{fig: heads}
\end{figure}

\vspace{-0.5em}
Point Head is applied for the localization task to track center points and knot key points of actors on each frame in a video clip. 
Specifically, our Point Head consists of two main detectors, a Center-Point (CP) and a Knot-Point (KP) detector, as shown in Figure~\ref{fig: heads}.

\noindent{\textbf{Center-Point (CP) detector.}} 
In the CP detector, there are three types of outputs. 
The first type is a \textit{center point heatmap}, denoted as $\mathbf{O}^{\text{CP}}_h \in \mathbb{R}^{1\times \frac{H}{R} \times \frac{W}{R}}$, for localizing the center point of an actor in the current frame. 
Following~\cite{law2018cornernet,cao2016realtime,zhou2019objects}, we apply a Gaussian kernel on the ground-truths of generated heatmaps.
We define the pixel-wise logistic regression loss $\mathcal{L}^{\text{CP}}_{h}$ between center point heatmaps as: 
\begin{equation}
\mathcal{L}^{\text{CP}}_{h}=-\frac{1}{N} \sum_{j=1}^{1\times \frac{H}{R} \times \frac{W}{R}}\left\{\begin{array}{cl}
\left(1-\widehat{g}_j\right)^{\alpha} \log \left(\widehat{g}_j\right) & \text { if }g_j=1 \\
\begin{array}{c}
\left(1-g_j\right)^{\beta}\left(\widehat{g}_j\right)^{\alpha} \\
\log \left(1-\widehat{g}_j\right)
\end{array} & \text { otherwise }
\end{array}\right.
\end{equation}
where $N$ denotes the number of actors in the frame.
$g_j$ and $\widehat{g}_j$ are the ground truth and prediction of the center point for pixel $j$ on heatmap $\mathbf{O}^{\text{CP}}_h$.
Following~\cite{zhou2019objects,law2018cornernet}, we set $\alpha = 2$ and $\beta = 4$ in our
experiments. 
The second type of output is the \textit{shape} prediction of a detected actor, denoted as $\mathbf{O}^{\text{CP}}_s \in \mathbb{R}^{2\times \frac{H}{R} \times \frac{W}{R}}$, which determines the height and width of the detected actor. 
The shape prediction loss $\mathcal{L}^{\text{CP}}_{s}$ is computed as
\begin{equation}
    \mathcal{L}^{\text{CP}}_{s}=\frac{1}{N} \sum_{i=1}^{N}\left|\widehat{s}_{i}-s_{i}\right|
\end{equation}

\noindent 
where $s_i = (x_{1}^{(i)}-x_{2}^{(i)},y_{1}^{(i)} - y_{2}^{(i)})$ and $\left(x_{1}^{(i)}, y_{1}^{(i)}, x_{2}^{(i)}, y_{2}^{(i)}\right)$ is the bounding box for $i$th actor. 
We denote its prediction as $\widehat{s}_i$.
The third type of output is an \textit{offset} refinement along the height and width of the actor, denoted as $\mathbf{O}^{\text{CP}}_o\in\mathbb{R}^{2\times \frac{H}{R} \times \frac{W}{R}}$, to minimize the effect caused by strides. 
When information of the detected actor projects back to the original frame, we apply this offset along $x$ and $y$ axes to make our localization result more accurate.
We define the offset loss $\mathcal{L}^{\text{CP}}_{o}$ inside each actor as
\begin{equation}
\mathcal{L}^{\text{CP}}_{o}=\frac{1}{N} \sum^{N}_{i}\left|\widehat{o}_{i}-\left(\frac{p_i}{R}-\widehat{p}_i\right)\right|
\end{equation}
where $\widehat{o}_i$ and $\widehat{p}_{i}$ are the predictions of the offset and the center point for $i$th actor. We denote $p_{i}$ as the ground-truth of the center point and $R$ as the stride size. 

As a result, the CP detector loss $\mathcal{L}^{\text{CP}}_{loc}$ is given by
\begin{equation}
\mathcal{L}^{\text{CP}}_{loc}=\lambda^{\text{CP}}_{h}\mathcal{L}^{\text{CP}}_{h}+\lambda^{c}_{s} \mathcal{L}^{\text{CP}}_{s}+\lambda^{\text{CP}}_{o}\mathcal{L}^{\text{CP}}_{o},
\end{equation}
where $\lambda^{\text{CP}}_{h}, \lambda^{\text{CP}}_{s}, \lambda^{\text{CP}}_{o}$ denote the weight hyper-parameter for the heatmap, shape and offset loss. 
We set $\lambda^{\text{CP}}_{h}=1, \lambda^{\text{CP}}_{s}=0.1, \lambda^{\text{CP}}_{o}=1$ in our all experiments.

\noindent{\textbf{Knot-Point (KP) detector.}} 
To make the action localization more accurate, we generate three types of outputs from the KP detector. 
The first output includes $K$-dimensional \textit{knot point heatmap}, denoted as $\mathbf{O}^{\text{KP}}_h \in \mathbb{R}^{K\times \frac{H}{R} \times \frac{W}{R}}$, which localizes $K$ knot points of the detected actor.  
The second output is $K$ \textit{distances} of knot points from the center point along the $x$ and $y$ axes, denoted as $\mathbf{O}^{\text{KP}}_d \in \mathbb{R}^{ 2K\times \frac{H}{R} \times \frac{W}{R} }$. 
The third output is an \textit{offset} refinement along the height and width of each key point, denoted as $\mathbf{O}^{\text{KP}}_o \in \mathbb{R}^{2\times \frac{H}{R} \times \frac{W}{R}}$. 
For the KP detector losses $L^{\text{KP}}_{h}$, we have similar loss calculation as the CP detector and the only difference is that we need to add the loss of $K$ points together. 
The the KP detector loss $\mathcal{L}^{\text{KP}}_{loc}$ is computed as
\begin{equation}
\mathcal{L}^{\text{KP}}_{loc}= \lambda^{\text{KP}}_{h}L^{\text{KP}}_{h}+ \lambda^{\text{KP}}_{d}L^{\text{KP}}_{d}+\lambda^{\text{KP}}_{o}L^{\text{KP}}_{o}
\end{equation}
where $\lambda^{\text{KP}}_{h}$, $\lambda^{\text{KP}}_{d}$ and $\lambda^{\text{KP}}_{o}$ denote the weight hyper-parameter for the knot point heatmap, distance and offset loss. 
We set $\lambda^{\text{KP}}_{h}=1, \lambda^{\text{KP}}_{d}=1, \lambda^{\text{KP}}_{o}=1$ in our all experiments.

\vspace{-1em}
\begin{figure}[!htb]
\setlength{\abovecaptionskip}{-1.5em}
\setlength{\belowcaptionskip}{-1em}
		\centerline{\includegraphics[width=0.65\textwidth]{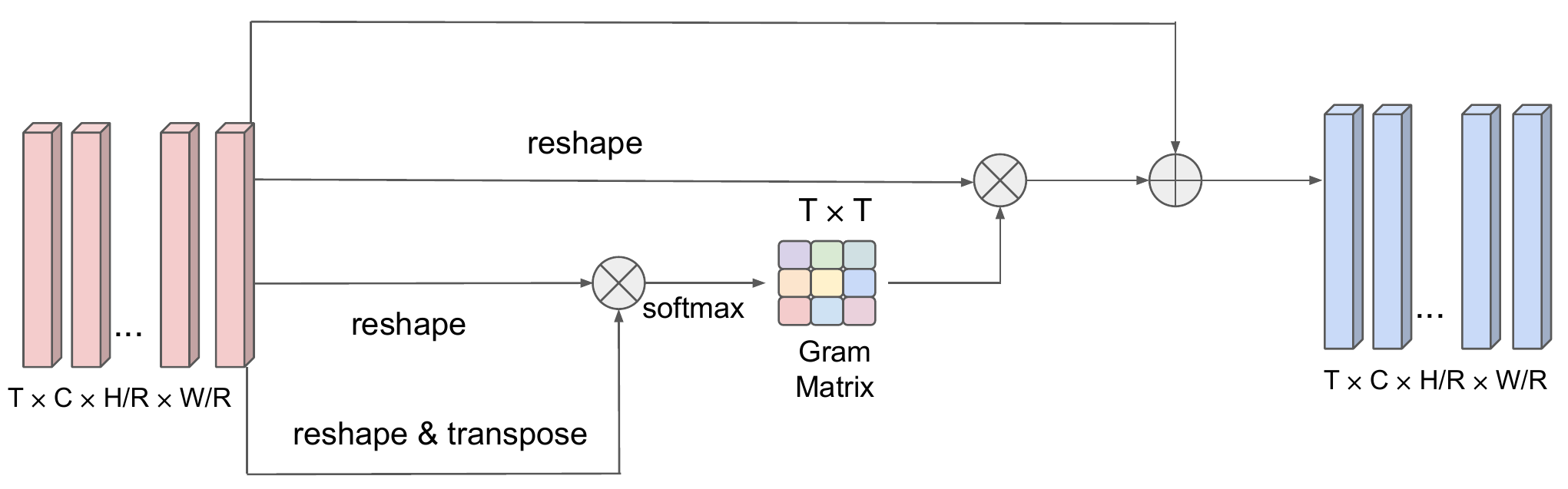}}
    	\caption{Time-wise Attention (TWA) mechanism for modeling dependencies across frames.}
	\label{fig: time att}
\end{figure}

\vspace{-0.5em}
\subsection{Time-wise Attention (TWA)}
\vspace{-0.5em}
In order to capture the long-range dependencies across frames, we introduce a Time-wise Attention module in our Point3D, as shown in Figure~\ref{fig: time att}. 
Specifically, we first stack features $\mathbf{f} \in \mathbb{R}^{C\times \frac{H}{R}\times \frac{W}{R}}$ of $T$ timestamps generated from Feature Extractor together as $\mathbf{F} \in \mathbb{R}^{T\times C\times \frac{H}{R}\times \frac{W}{R}}$ before feeding them into TWA module. 
Then, we reshape $\mathbf{F}$ into $\Tilde{\mathbf{F}} \in \mathbb{R}^{T\times (C\times \frac{H}{R}\times \frac{W}{R})}$. 
We take the matrix multiplication between this reshaped matrix $\Tilde{\mathbf{F}}$ and its transpose to get the Gram matrix $\mathbf{G}$, which indicates the correlations among each time step.
The Gram matrix is then passed through a softmax layer and get the time-wise attention $M$, where each entry inside $M$ measures how one frame is related to another. 
To avoid gradient vanishing on the original feature map, $\Tilde{\mathbf{F}}$ is multiplied by the Gram matrix $\mathbf{G}$ to generate matrix $\widehat{\mathbf{S}}$.
Finally, we reshape $\widehat{\mathbf{S}}$ back to the shape of original matrix as $\mathbf{S}$ and add the original matrix $\mathbf{F}$ to get output $\mathbf{Y}$.
In this way, temporal information between frames has been enhanced to help the 3D Head for later action classification. 

\vspace{-1em}
\begin{figure}[!htb]
\setlength{\abovecaptionskip}{-1.5em}
\setlength{\belowcaptionskip}{-1em}
		\centering
		\includegraphics[width=0.55\linewidth]{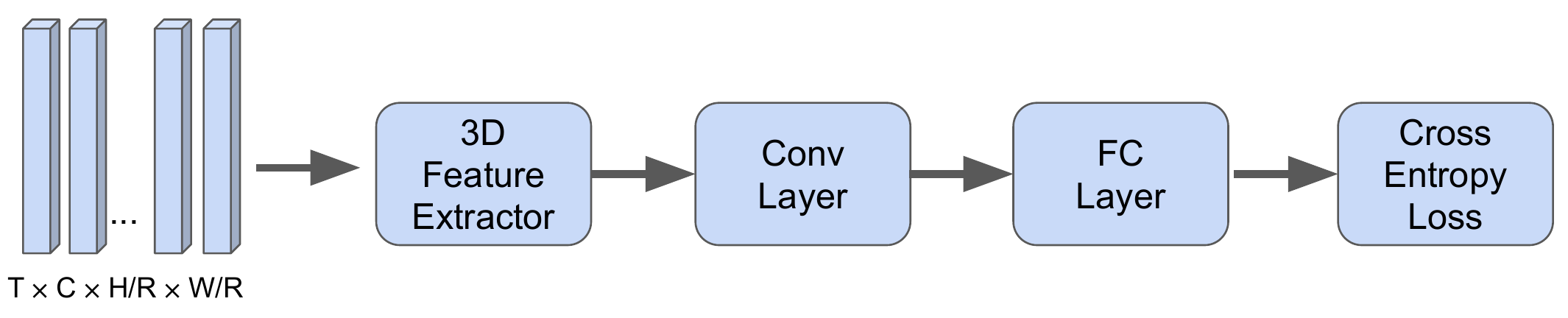}
    	\caption{\textbf{Pipeline of 3D head.} The output from TWA are \red{fed} into a 3D CNN backbone to extract the features, then we input these features to two convolutional layers and one fully-connected layer to do the class prediction. Finally, a cross entropy loss is applied to the output for better classification.}
	\label{fig: 3D CNN}
\end{figure}

\vspace{-1.5em}
\subsection{3D Head}
\vspace{-0.5em}
To improve the performance of action classification, we introduce 3D CNNs in our 3D Head in Figure~\ref{fig: 3D CNN}.  
Concretely, we first feed the output $\mathbf{Y}$ from TWA into a 3D backbone as the feature extractor for its robust performance~\cite{Xie2016}. 
Then, we pipe these generated 3D features into two convolutional layers followed by a fully-connected layer and obtain action label index. 
During the training process, cross entropy loss is used to train the 3D Head. 
Assume $D$ is the one-hot encoding vector of ground truth class label.
We define the classification loss $\mathcal{L}_{cls}$ on the output of 3D Head as
\begin{equation}
    \mathcal{L}_{cls} = -\log(\widehat{d})
\end{equation}
where $\widehat{d}$ be the softmax output for the correct class entry in $D$.

Putting localization and classification loss together, the overall loss $\mathcal{L}_{overall}$ of our Point3D is simply computed as 
\begin{equation}
    \mathcal{L}_{overall} = \lambda_{loc} \mathcal{L}_{loc} + \lambda_{cls} \mathcal{L}_{cls}, 
\end{equation}
where $\lambda_{loc}, \lambda_{cls}$ denote the weight hyper-parameter of localization and classification loss. 
In order to explore how much each loss affects the final performance of our Point3D, we perform extensive experiments on each parameter in the supplementary.  


\noindent{\textbf{Tubelet linking.}}
After obtaining the bounding box on each frame, we need to link each frame together into the video clip. 
We use the linking algorithm from~\cite{gkioxari2015finding} and compute scores with the definition in~\cite{kopuklu2019yowo}. 
After all the linking scores are computed, the Viterbi algorithm~\cite{forney1973viterbi} is applied to find the optimal action tube. 
\vspace{-1em}
\section{Experiments}
\vspace{-0.5em}

\subsection{Experiments Settings}
\vspace{-0.5em}
\noindent\textbf{Datasets and Metrics.} 
We perform experiments on the UCF101-24~\cite{khurram2012ucf}, JHMDB~\cite{Jhuang:ICCV:2013}, and AVA~\cite{gu2018ava} datasets. 
UCF101-24 consists of 3207 temporally untrimmed videos from 24 sports classes~\cite{khurram2012ucf}. 
We report the action detection performance for the first split only following the common settings as in~\cite{li2020actions, kopuklu2019yowo,kalogeiton2017action,peng2016multi}. 
JHMDB is a subset of the HMDB-51 dataset and consists of 928 short videos with 21 action categories in daily life~\cite{Jhuang:ICCV:2013}.
Each video of JHMDB is well trimmed and has a single action instance across all the frames. 
We report results averaged over three splits following the common settings as in~\cite{li2020actions, kopuklu2019yowo,kalogeiton2017action,peng2016multi}.  
Following state-of-the-art methods~\cite{li2020actions,kopuklu2019yowo,weinzaepfel2015learning,gkioxari2015finding,kalogeiton2017action}, we adopt frame mAP and video mAP to evaluate action detection accuracy.
AVA is a more challenging dataset consisting of 211k training and 57k validation video segments, where each actor in the key frame is labeled at 1 FPS and with one bounding box and multiple classes from 80 atomic action categories. 
Following the common protocol~\cite{tang2020asynchronous,kopuklu2019yowo,gu2018ava}, we report frame mAP performance with an IoU threshold of 0.5 over the top 60 action classes on AVA v2.1 and v2.2 benchmarks.

\noindent\textbf{Implementation Details.} 
We employ the Hourglass-104~\cite{alejandro2016hourglass} as the feature extractor network, 
closely following the same setting as the current state-of-the-art methods~\cite{li2020actions,kopuklu2019yowo} for a fair comparison. 
The frame is resized to $256\times 256$ for effective computations. 
The spatial down-sampling ratio $R$ is set to $4$ and the resulted feature map size is $64\times64$. 
The number of input frames $T$ is 16. 
We use Adam with a learning rate of 5e-4 to optimize the feature extractor and the Point Head. 
The 3D Head is optimized by using a SGD with a learning rate of 1e-3. 
The overall network is trained in an end-to-end manner. 
The learning rate adjusts to convergence on the validation set and decreases by a factor of $10$ when performance saturates. 
For a fair comparison, we closely follow previous work~\cite{li2020actions,tang2020asynchronous,wu2020contextaware} to train $12$ epochs on UCF101-24~\cite{khurram2012ucf}, $20$ epochs on JHMDB~\cite{Jhuang:ICCV:2013}, and $15$ epochs on AVA~\cite{gu2018ava}. The total training time on 8 Tesla V100 GPUs is 153 hours.

\vspace{-1em}
\subsection{Comparison with State of the Arts}
\vspace{-0.5em}

\begin{table}[!htb]
	\caption{Comparison with state-of-the-art methods on the JHMDB and UCF101-24 benchmark. Bold and underlined numbers denote the first and second place.}
	\label{tab: sota_jhmdb_ucf}
	\renewcommand\tabcolsep{6.0pt}
	\centering
	\scalebox{0.65}{
\begin{tabular}{lccccccccccc}
\toprule
\multicolumn{1}{l}{\multirow{3}{*}{Method}}& \multicolumn{5}{c}{JHMDB} & \multicolumn{6}{c}{UCF101-24} \\
& \multicolumn{1}{c}{Frame-mAP} & \multicolumn{4}{c}{Video-mAP(\%)}  & \multicolumn{1}{c}{Frame-mAP} & \multicolumn{5}{c}{Video-mAP(\%)}                   \\  
\multicolumn{1}{c}{}                        & \multicolumn{1}{c}{0.5(\%)}                                & 0.2  & 0.5  & 0.75 & \multicolumn{1}{c}{0.5:0.95} & \multicolumn{1}{c}{0.5(\%)}                                & 0.1 & 0.2  & 0.5  & 0.75 & \multicolumn{1}{c}{0.5:0.95} \\ \midrule
\multicolumn{2}{l}{\textit{2D backbone:}}                                \\ 
MR-TS R-CNN~\cite{peng2016multi}                                          & 58.5                                                 & 74.3 & 73.1 &  -   &  -                          & 39.9                            & 50.4   & 42.3  &  -   &  -   &  - \\ 
A+AF~\cite{singh2017online}                                         & -                                                    & 73.8 & 72   & 44.5 & 41.6                          & -                                &  -      & 73.5  & 46.3  & 15.0  & 20.4     \\ 
SSD+ACT~\cite{kalogeiton2017action}                                    & 65.7                                                 & 74.2 & 73.7 & 52.1 & 44.8                          & 69.5                            &   -     & 76.5  & 49.2  & 19.7  & 23.4     \\  
TACNet~\cite{song2019tacnet}                                          & 65.5                                                 & 74.1 & 73.4 & 52.5 & 44.8                          & 72.1                            &   -     & 77.5  & 52.9  & 21.8  & 24.1     \\
Dance+Flow~\cite{zhao2019dance}                                          & -                                                   &  -   & 74.7 & 53.3 & 45.0 & -                                &   -     & 78.5  & 50.3  & 22.2  & 24.5     \\
MOC~\cite{li2020actions}                                         & 70.8                                                 & 77.3 & 77.2 & \textbf{71.7} & 59.1                         & 78.0                            &   -     & \underline{82.8}  & 53.8  & \underline{29.6}  & \underline{28.3}     \\ \hline
\multicolumn{2}{l}{\textit{3D backbone:}}                             \\
C3D \cite{hou2017tube}                                           & 61.3                                                 & 78.4 & 76.9 &  -   &  -                           & 41.4                            &   -     & 47.1  &  -  &  -   & -     \\ 
I3D~\cite{gu2018ava}                                           & 73.3                                                 &  -   & 78.6 &  -   &  -                           & 76.3                            &   -     &  -   & \textbf{59.9}  &  -  & -       \\
S3D-G \cite{sun2018actor}                                         & \underline{77.9}                                                 &  -   & 80.1 &  -   &  -      &  -   &  - &  -   &  - &  -   &  -                       \\
\hline
\multicolumn{2}{l}{\textit{2D+3D backbone:}} \\ 
YOWO \cite{kopuklu2019yowo}                                          & 74.4                                                 & \underline{87.8} & \underline{85.7} & 58.1 &  -   & \underline{80.4}                            & 82.5   & 75.8  & 48.8  &   -   &  -                               \\ 
Point3D (ours)                                &                 \textbf{79.2} & \textbf{89.1} & \textbf{86.1} & \underline{71.5} & \textbf{60.9}      &      \textbf{83.5} & \textbf{85.4} & \textbf{84.5} & \underline{55.1} & \textbf{33.4} & \textbf{31.8} \\ \bottomrule
\end{tabular}}
\end{table}

\noindent{\textbf{JHMDB.}} We compare our method to the current state-of-the-art methods, including $2$D backbone methods (SSD+ACT~\cite{kalogeiton2017action}, TACNet~\cite{song2019tacnet}, Dance+Flow~\cite{zhao2019dance}, MOC~\cite{li2020actions}) and 3D backbone methods (C3D \cite{hou2017tube}, I3D \cite{gu2018ava}, S3D-G \cite{sun2018actor}). 
Table~\ref{tab: sota_jhmdb_ucf} shows the results on the JHMDB benchmark. 
Our method outperforms the state-of-the-art methods by a large margin in terms of both frame-mAP and three video-mAP metrics. Except for video-mAP with a threshold of 0.75 where we achieve a comparable result as MOC~\cite{li2020actions}, our Point3D outperforms all other methods. 

\noindent{\textbf{UCF101-24.}}
We compare our method with the state-of-the-art methods in Table~\ref{tab: sota_jhmdb_ucf}. 
Our Point3D outperforms the current state-of-the-art methods by a significant margin over all metrics. 
This direct comparisons demonstrate the improvement using Point3D. Introducing the 3D Head in Point3D helps to outperform the 2D detectors~\cite{peng2016multi, singh2017online, kalogeiton2017action, song2019tacnet, zhao2019dance, li2020actions} by a large margin. 
We also achieve a better performance than YOWO \cite{kopuklu2019yowo} with the 3D detector with the help of our Point Head that tracks actions as moving points.

\noindent{\textbf{AVA.}} 
Following previous work~\cite{kopuklu2019yowo,tang2020asynchronous,wu2020contextaware}, we report comparison results on the AVA v2.1 and v2.2 benchmarks in Table~\ref{tab: sota_ava}.
As we can see, our Point3D pre-trained on Kinetics-400~\cite{kay2017kinetics} outperforms YOWO~\cite{kopuklu2019yowo} by 10.3\% mAP, which verifies the effectiveness of the proposed anchor-free architecture in spatio-temporal action localization. 
Typically, our Point3D pre-trained on Kinetics-700~\cite{kay2017kinetics} achieves a new state-of-the-art performance on both benchmarks. 
We also achieve comparable results as AIA~\cite{tang2020asynchronous}, although we did not leverage different interactions to improve action detection performance.

\vspace{-0.5em}
\begin{table}[!htb]
	\caption{Comparison with state-of-the-art methods on the AVA v2.1 and v2.2 benchmark.}
	\label{tab: sota_ava}
	\renewcommand\tabcolsep{6.0pt}
	\centering
	\scalebox{0.65}{
\begin{tabular}{lccc}
\toprule
Method & Pretrain & v2.1 & v2.2 \\ \midrule
\multicolumn{2}{l}{\textit{2D backbone:}}                           \\ 
Relation Graph~\cite{zhang2019a}	& Kinetics-400 & 22.2 & -\\
LFB~\cite{wu2019long}	& Kinetics-400 & 27.7 &  - \\
SlowFast~\cite{christoph2019slowfast}	& Kinetics-600 & 28.2 & 29.1 \\
              \hline
\multicolumn{2}{l}{\textit{3D backbone:}}                             \\ 
I3D~\cite{gu2018ava} & Kinetics-400 & 15.6 & -\\
S3D~\cite{sun2018actor} & Kinetics-400 & 17.4 & -\\
VAT~\cite{girdhar2019video}	& Kinetics-400 & 25.0  & - \\  
C-A RCNN~\cite{wu2020contextaware}	& Kinetics-400 & 28.0 & -\\    
AIA~\cite{tang2020asynchronous} &Kinetics-700 & 31.2 & 32.3 \\
\hline
\multicolumn{2}{l}{\textit{2D+3D backbone:}}      \\ 
YOWO~\cite{kopuklu2019yowo} & Kinetics-400 & 19.2 & 20.2 \\
Point3D (ours) & Kinetics-400 & 29.5 & 30.6\\
Point3D (ours) & Kinetics-700 & \textbf{31.6} & \textbf{32.8}\\ \bottomrule
\end{tabular}}
\end{table}

\begin{figure}[!htb]
\setlength{\abovecaptionskip}{-1.5em}
\setlength{\belowcaptionskip}{-1em}
		\centerline{\includegraphics[width=0.6\linewidth]{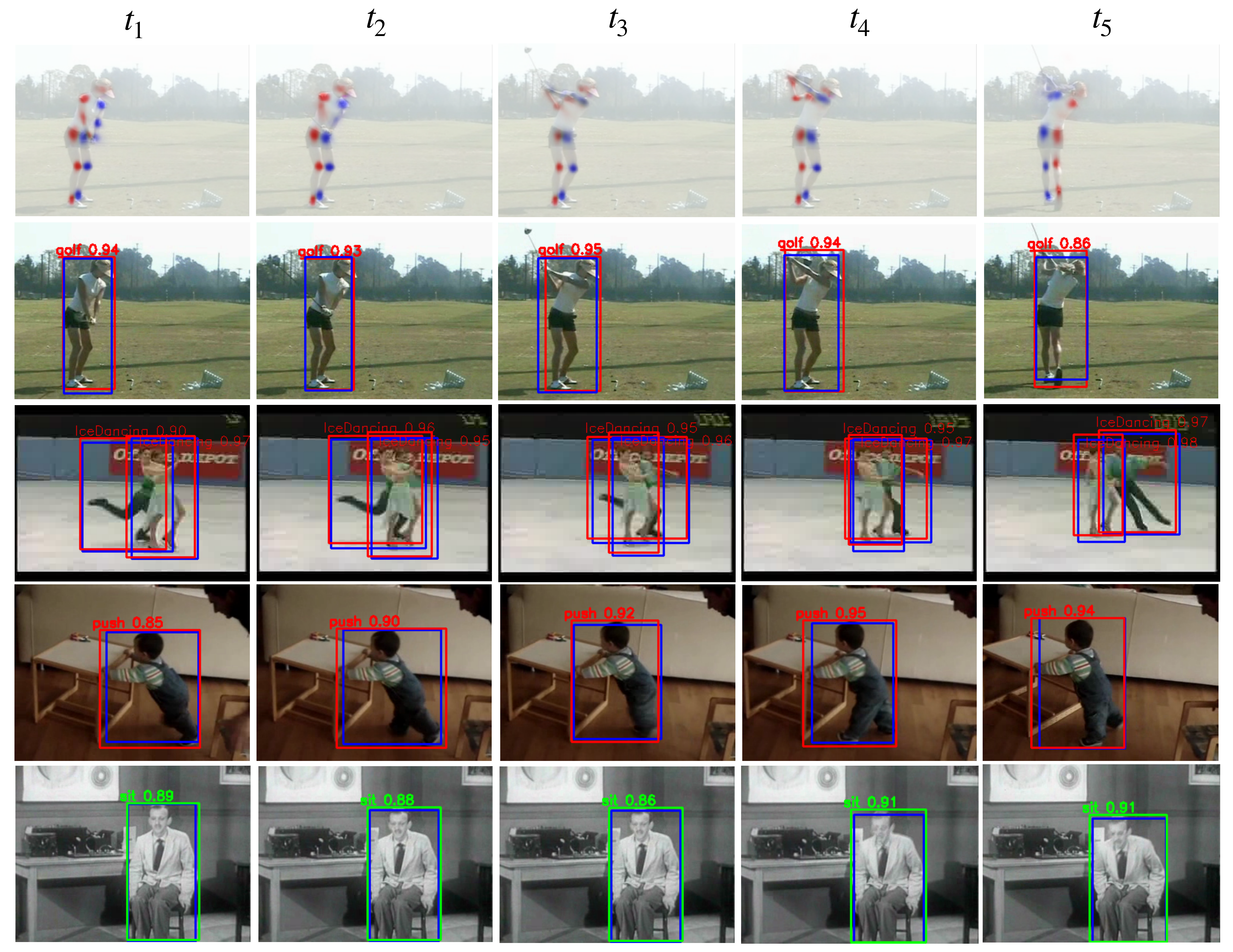}}
    	\caption{Visualization results. The \textbf{first row} denotes the heatmaps generated from our KP detector. The \textbf{blue} bounding boxes
are ground truth while \textbf{red} and \textbf{green} are true and false positive detections, respectively.  Zoom in for a better view.}
\vspace{-0.5em}
	\label{fig: exp_visualization}
\end{figure}

\vspace{-0.5em}
\noindent{\textbf{Visualizations.}}
In Figure \ref{fig: exp_visualization}, we visualize some qualitative examples of action detection on JHMDB and UCF101-24 datasets. In general, our Point3D architecture exhibits a decent job at localizing actions in videos. As can be seen in the heatmaps on the top row, our Point3D get rid of the anchor to localize and classify the action by tracking the spatial and temporal changes of key points in the frames.
We defer more visualization examples to the supplementary materials.
We also visualize failure cases in the last row in Figure~\ref{fig: exp_visualization}. 
Our Point3D sometimes makes some false positive classifications at initial frames since it is hard to recognize the action before it happens.
\vspace{-1em}

\vspace{-0.5em}
\subsection{Ablation Study}\label{sec: ablation_study}
\vspace{-0.5em}
In this section, we explore extensive ablation studies on each part of our Point3D, including the Point Head, the backbone and the input of the 3D Head, the temporal length of input clips, and the hyper-parameters for weighting the overall loss. Unless specified, we conduct all ablation studies on the JHMDB benchmark.

\noindent{\textbf{Each component of Point3D.}}
In Table \ref{tab: ab_cent}, we explore each part of our Point3D, including CP detector, KP detector, 3D Head, and TWA. 
Introducing KP in the CP and CP+3D Head increase the frame-mAP by 0.4\% and 1.3\%, respectively. Adding 3D Head to CP and CP+KP improve the frame-mAP by 6.7\% and 7.6\%, respectively. This demonstrates the advantage of 3D Head in action classification. Adding TWA to CP+KP and CP+KP+3D Head further increases the frame-mAP by 2.0\% and 1.1\%, respectively, which shows the validity of the proposed Time-wise Attention in our Point3D. Similar improvements are also shown in the four video-mAP metrics.


\begin{table}[!htb]
	\caption{Exploration study on each part of our Point3D.}
	\label{tab: ab_cent}
	\renewcommand\tabcolsep{6.0pt}
	\centering
	\scalebox{0.65}{
		\begin{tabular}{ccccccccc}
			\toprule 
			\multicolumn{1}{c}{\multirow{2}{*}{CP}} & \multicolumn{1}{c}{\multirow{2}{*}{KP}} & 
			\multicolumn{1}{c}{3D} & 
			\multicolumn{1}{c}{\multirow{2}{*}{TWA}} & \multicolumn{1}{c}{Frame-mAP(\%)} & \multicolumn{4}{c}{Video-mAP(\%)}                     \\ 
   \multicolumn{1}{c}{}& \multicolumn{1}{c}{}&\multicolumn{1}{c}{Head} &\multicolumn{1}{c}{}  & \multicolumn{1}{c}{0.5}                                & \multicolumn{1}{c}{0.2}  & \multicolumn{1}{c}{0.5}  & \multicolumn{1}{c}{0.75} & \multicolumn{1}{c}{0.5:0.95} \\ \midrule
            \checkmark& & & & 70.1 & 76.5 & 77.2 & 69.1 & 58.3 \\
			\checkmark & \checkmark & & & 70.5 & 77.8 & 78.3 & 69.9 & 58.5 \\
		    \checkmark & \checkmark &  &\checkmark & 72.5 & 79.3 & 80.1 & 70.4 & 59.3 \\
		     \checkmark& &\checkmark &  & 76.8 & 85.6 & 82.7 & 70.7 & 59.8 \\
		     \checkmark  & \checkmark & \checkmark& & 78.1 & 88.2 & 85.6 & 70.8 & 60.4 \\
		     \checkmark  & \checkmark & \checkmark& \checkmark &\textbf{79.2} & \textbf{89.1} & \textbf{86.1} & \textbf{71.5} & \textbf{60.9} \\
			\bottomrule
			\end{tabular}}
\end{table}




\vspace{-1em}
\begin{table}[!htb]
	\caption{Exploration study on the temporal length $T$ of input clips.}
	\label{tab: ab_duration}
	\renewcommand\tabcolsep{6.0pt}
	\centering
	\scalebox{0.65}{
		\begin{tabular}{ccccccc}
			\toprule
			\multicolumn{1}{c}{\multirow{2}{*}{$T$}} & \multicolumn{1}{c}{\multirow{2}{*}{Speed(fps)}} & \multicolumn{1}{c}{Frame-mAP(\%)} & \multicolumn{4}{c}{Video-mAP(\%)}                     \\ 
\multicolumn{1}{c}{}    & \multicolumn{1}{c}{}   & \multicolumn{1}{c}{0.5}                                & 0.2  & 0.5  & 0.75 & \multicolumn{1}{c}{0.5:0.95} \\ \midrule
            1 & 38 & 69.8 & 78.5 & 77.4 & 63.2 & 53.8 \\
            3 & 32 &  72.1 & 81.4 & 80.8 & 66.4 & 56.7 \\
            5 & 27 &  76.3 & 85.2 & 83.6 & 69.5 & 59.6 \\
            7 & 23 &  78.6 & 88.5 & 85.4 & 70.8 & 60.3 \\
            9 &   19 & 78.8 & 88.7 & 85.8 & 71.1 & 60.5 \\
			13 & 15 & 78.9 & 88.9 & 85.9 & 71.2 & 60.7 \\
			16  & 13 & 79.2 & 89.1 & 86.1 & 71.5 & 60.9 \\
			32 & 7 & \textbf{79.8} & \textbf{89.6} & \textbf{86.5} & \textbf{71.9} & \textbf{61.4} \\
			\bottomrule
			\end{tabular}}
\end{table}

\noindent{\textbf{Temporal Length of Input Clip.}}
In this study, we explore the temporal length $T$ of the input clip in our Point3D. We vary $T$ from $1$ to $32$ and report the results in Table \ref{tab: ab_duration}.
First, we observe that when $T=7$, our Point3D outperforms the single-frame detector by a large margin in terms of frame-mAP and video-mAP metrics.
This confirms the importance of temporal information for action recognition, which agrees with the motivation of adding a 3D Head to our Point3D.
Second, we note that the detection performance will increase as we vary $T$ from $1$ to $32$ and the performance increase scope becomes smaller.
Third, when $T=32$, we follow~\cite{kopuklu2019yowo} and introduce the long-term feature bank~\cite{wu2019long} in the input of 3D Head, which brings further improvements on the performance.
\vspace{-0.5em}


\vspace{-0.2cm}
\section{Conclusion}
In this paper, we propose Point3D, a novel anchor-free architecture that is flexible in structure and enjoy good computation efficiency for the task of spatio-temporal action localization. Point3D consists of a \textit{Point Head} for action localization and a \textit{3D Head} for action classification.
We conduct extensive experiments on the JHMDB, UCF101-24 and AVA benchmarks where our method achieves the state-of-the-art results in terms of frame-mAP and video-mAP.
Comprehensive ablation studies also demonstrate the effectiveness of the Point Head, the 3D Head, and TWA proposed in our Point3D.

\section*{Acknowledgement}

We thank Pinxu Ren for insightful discussions.


\bibliography{reference}

\begin{thebibliography}{50}
\providecommand{\natexlab}[1]{#1}
\providecommand{\url}[1]{\texttt{#1}}
\expandafter\ifx\csname urlstyle\endcsname\relax
  \providecommand{\doi}[1]{doi: #1}\else
  \providecommand{\doi}{doi: \begingroup \urlstyle{rm}\Url}\fi

\bibitem[Bochkovskiy et~al.(2020)Bochkovskiy, Wang, and
  Liao]{bochkovskiy2020yolov4}
Alexey Bochkovskiy, Chien-Yao Wang, and Hong-Yuan~Mark Liao.
\newblock {YOLOv4:} optimal speed and accuracy of object detection.
\newblock \emph{arXiv preprint arXiv:2004.10934}, 2020.

\bibitem[Cao et~al.(2016)Cao, Simon, Wei, and Sheikh]{cao2016realtime}
Zhe Cao, Tomas Simon, Shih{-}En Wei, and Yaser Sheikh.
\newblock Realtime multi-person 2d pose estimation using part affinity fields.
\newblock In \emph{Proceedings of the IEEE/CVF Conference on Computer Vision
  and Pattern Recognition}, pages 7291--7299, 2016.

\bibitem[Coppola et~al.(2016)Coppola, Faria, Nunes, and
  Bellotto]{coppola2016social}
Claudio Coppola, Diego~R Faria, Urbano Nunes, and Nicola Bellotto.
\newblock Social activity recognition based on probabilistic merging of
  skeleton features with proximity priors from rgb-d data.
\newblock In \emph{Proceedings of the IEEE/RSJ International Conference on
  Intelligent Robots and Systems}, pages 5055--5061, 2016.

\bibitem[Coppola et~al.(2019)Coppola, Cosar, Faria, and
  Bellotto]{coppola2019social}
Claudio Coppola, Serhan Cosar, Diego~R Faria, and Nicola Bellotto.
\newblock Social activity recognition on continuous rgb-d video sequences.
\newblock \emph{International Journal of Social Robotics}, pages 1--15, 2019.

\bibitem[Duan et~al.(2019)Duan, Bai, Xie, Qi, Huang, and
  Tian]{duan2019centernet}
Kaiwen Duan, Song Bai, Lingxi Xie, Honggang Qi, Qingming Huang, and Qi~Tian.
\newblock Centernet: Keypoint triplets for object detection.
\newblock In \emph{Proceedings of the IEEE/CVF International Conference on
  Computer Vision}, pages 6569--6578, 2019.

\bibitem[Duarte et~al.(2018)Duarte, Rawat, and Shah]{duarte2018videocapsulenet}
Kevin Duarte, Yogesh Rawat, and Mubarak Shah.
\newblock Videocapsulenet: A simplified network for action detection.
\newblock In \emph{Proceedings of the Advances in Neural Information Processing
  Systems}, 2018.

\bibitem[Feichtenhofer et~al.(2019)Feichtenhofer, Fan, Malik, and
  He]{christoph2019slowfast}
Christoph Feichtenhofer, Haoqi Fan, Jitendra Malik, and Kaiming He.
\newblock Slowfast networks for video recognition.
\newblock In \emph{Proceedings of the IEEE/CVF International Conference on
  Computer Vision}, pages 6202--6211, 2019.

\bibitem[Forney(1973)]{forney1973viterbi}
G~David Forney.
\newblock The viterbi algorithm.
\newblock \emph{Proceedings of the the IEEE}, 61\penalty0 (3):\penalty0
  268--278, 1973.

\bibitem[Girdhar et~al.(2019)Girdhar, Carreira, Doersch, and
  Zisserman]{girdhar2019video}
Rohit Girdhar, Joao Carreira, Carl Doersch, and Andrew Zisserman.
\newblock Video action transformer network.
\newblock In \emph{Proceedings of the IEEE/CVF Conference on Computer Vision
  and Pattern Recognition}, pages 244--253, 2019.

\bibitem[Girshick(2015)]{girshick2015fast}
Ross Girshick.
\newblock {Fast R-CNN}.
\newblock In \emph{Proceedings of the IEEE/CVF International Conference on
  Computer Vision}, pages 1440--1448, 2015.

\bibitem[Girshick et~al.(2014)Girshick, Donahue, Darrell, and
  Malik]{girshick2014rich}
Ross Girshick, Jeff Donahue, Trevor Darrell, and Jitendra Malik.
\newblock Rich feature hierarchies for accurate object detection and semantic
  segmentation.
\newblock In \emph{Proceedings of the IEEE/CVF conference on Computer Vision
  and Pattern Recognition}, pages 580--587, 2014.

\bibitem[Gkioxari and Malik(2015)]{gkioxari2015finding}
Georgia Gkioxari and Jitendra Malik.
\newblock Finding action tubes.
\newblock In \emph{Proceedings of the IEEE/CVF Conference on Computer Vision
  and Pattern Recognition}, pages 759--768, 2015.

\bibitem[Gu et~al.(2018)Gu, Sun, Ross, Vondrick, Pantofaru, Li,
  Vijayanarasimhan, Toderici, Ricco, Sukthankar, Schmid, and Malik]{gu2018ava}
Chunhui Gu, Chen Sun, David~A. Ross, Carl Vondrick, Caroline Pantofaru, Yeqing
  Li, Sudheendra Vijayanarasimhan, George Toderici, Susanna Ricco, Rahul
  Sukthankar, Cordelia Schmid, and Jitendra Malik.
\newblock {AVA:} a video dataset of spatio-temporally localized atomic visual
  actions.
\newblock In \emph{Proceedings of the IEEE/CVF Conference on Computer Vision
  and Pattern Recognition}, pages 6047--6056, 2018.

\bibitem[He et~al.(2018)He, Deng, Ibrahim, and Mori]{he2017generic}
Jiawei He, Zhiwei Deng, Mostafa~S. Ibrahim, and Greg Mori.
\newblock Generic tubelet proposals for action localization.
\newblock In \emph{Proceedings of the IEEE/CVF Winter Conference on
  Applications of Computer Vision}, pages 343--351, 2018.

\bibitem[Hou et~al.(2017)Hou, Chen, and Shah]{hou2017tube}
Rui Hou, Chen Chen, and Mubarak Shah.
\newblock Tube convolutional neural network (t-cnn) for action detection in
  videos.
\newblock In \emph{Proceedings of the IEEE/CVF International Conference on
  Computer Vision}, pages 5822--5831, 2017.

\bibitem[Jhuang et~al.(2013)Jhuang, Gall, Zuffi, Schmid, and
  Black]{Jhuang:ICCV:2013}
Hueihan Jhuang, Juergen Gall, Silvia Zuffi, Cordelia Schmid, and Michael~J.
  Black.
\newblock Towards understanding action recognition.
\newblock In \emph{Proceedings of the IEEE/CVF International Conference on
  Computer Vision}, pages 3192--3199, 2013.

\bibitem[Kalogeiton et~al.(2017)Kalogeiton, Weinzaepfel, Ferrari, and
  Schmid]{kalogeiton2017action}
Vicky Kalogeiton, Philippe Weinzaepfel, Vittorio Ferrari, and Cordelia Schmid.
\newblock Action tubelet detector for spatio-temporal action localization.
\newblock In \emph{Proceedings of the IEEE/CVF International Conference on
  Computer Vision}, pages 4405--4413, 2017.

\bibitem[Kay et~al.(2017)Kay, Carreira, Simonyan, Zhang, Hillier,
  Vijayanarasimhan, Viola, Green, Back, Natsev, et~al.]{kay2017kinetics}
Will Kay, Joao Carreira, Karen Simonyan, Brian Zhang, Chloe Hillier, Sudheendra
  Vijayanarasimhan, Fabio Viola, Tim Green, Trevor Back, Paul Natsev, et~al.
\newblock The kinetics human action video dataset.
\newblock \emph{arXiv preprint arXiv:1705.06950}, 2017.

\bibitem[K{\"o}p{\"u}kl{\"u} et~al.(2019)K{\"o}p{\"u}kl{\"u}, Wei, and
  Rigoll]{kopuklu2019yowo}
Okan K{\"o}p{\"u}kl{\"u}, Xiangyu Wei, and Gerhard Rigoll.
\newblock You only watch once: A unified cnn architecture for real-time
  spatiotemporal action localization.
\newblock \emph{arXiv preprint arXiv:1911.06644}, 2019.

\bibitem[Law and Deng(2018)]{law2018cornernet}
Hei Law and Jia Deng.
\newblock Cornernet: Detecting objects as paired keypoints.
\newblock In \emph{Proceedings of the European Conference on Computer Vision},
  pages 734--750, 2018.

\bibitem[Leo et~al.(2004)Leo, D'Orazio, and Spagnolo]{leo2004human}
Marco Leo, Tiziana D'Orazio, and Paolo Spagnolo.
\newblock Human activity recognition for automatic visual surveillance of wide
  areas.
\newblock In \emph{Proceedings of the ACM 2nd international workshop on Video
  surveillance \& sensor networks}, pages 124--130, 2004.

\bibitem[Li et~al.(2020)Li, Wang, Wang, and Wu]{li2020actions}
Yixuan Li, Zixu Wang, Limin Wang, and Gangshan Wu.
\newblock Actions as moving points.
\newblock In \emph{Proceedings of the European Conference on Computer Vision},
  pages 68--84, 2020.

\bibitem[Mitra and Acharya(2007)]{mitra2007gesture}
Sushmita Mitra and Tinku Acharya.
\newblock Gesture recognition: A survey.
\newblock \emph{IEEE Transactions on Systems, Man, and Cybernetics, Part C
  (Applications and Reviews)}, 37\penalty0 (3):\penalty0 311--324, 2007.

\bibitem[Mo et~al.(2020)Mo, Tan, Xia, and Ren]{mo2020improving}
Shentong Mo, Xiaoqing Tan, Jingfei Xia, and Pinxu Ren.
\newblock Towards improving spatiotemporal action recognition in videos.
\newblock \emph{arXiv preprint arXiv:2012.08097}, 2020.

\bibitem[Newell et~al.(2016)Newell, Yang, and Deng]{alejandro2016hourglass}
Alejandro Newell, Kaiyu Yang, and Jia Deng.
\newblock Stacked hourglass networks for human pose estimation.
\newblock In \emph{Proceedings of the European Conference on Computer Vision},
  pages 483--499, 2016.

\bibitem[Oh et~al.(2011)Oh, Hoogs, Perera, Cuntoor, Chen, Lee, Mukherjee,
  Aggarwal, Lee, Davis, et~al.]{oh2011large}
Sangmin Oh, Anthony Hoogs, Amitha Perera, Naresh Cuntoor, Chia-Chih Chen,
  Jong~Taek Lee, Saurajit Mukherjee, JK~Aggarwal, Hyungtae Lee, Larry Davis,
  et~al.
\newblock A large-scale benchmark dataset for event recognition in surveillance
  video.
\newblock In \emph{Proceedings of the IEEE/CVF Conference on Computer Vision
  and Pattern Recognition}, pages 3153--3160, 2011.

\bibitem[Peng and Schmid(2016)]{peng2016multi}
Xiaojiang Peng and Cordelia Schmid.
\newblock Multi-region two-stream r-cnn for action detection.
\newblock In \emph{Proceedings of the European Conference on Computer Vision},
  pages 744--759, 2016.

\bibitem[Rautaray and Agrawal(2015)]{rautaray2015vision}
Siddharth~S Rautaray and Anupam Agrawal.
\newblock Vision based hand gesture recognition for human computer interaction:
  a survey.
\newblock \emph{Artificial Intelligence Review}, 43\penalty0 (1):\penalty0
  1--54, 2015.

\bibitem[Redmon and Farhadi(2017)]{redmon2016yolo9000}
Joseph Redmon and Ali Farhadi.
\newblock {YOLO9000:} better, faster, stronger.
\newblock In \emph{Proceedings of the IEEE/CVF Conference on Computer Vision
  and Pattern Recognition}, pages 6517--6525, 2017.

\bibitem[Redmon and Farhadi(2018)]{redmon2018yolov3}
Joseph Redmon and Ali Farhadi.
\newblock {YOLOv3:} an incremental improvement.
\newblock \emph{arXiv preprint arXiv:1804.02767}, 2018.

\bibitem[Redmon et~al.(2016)Redmon, Divvala, Girshick, and
  Farhadi]{redmon2016you}
Joseph Redmon, Santosh Divvala, Ross Girshick, and Ali Farhadi.
\newblock You only look once: Unified, real-time object detection.
\newblock In \emph{Proceedings of the IEEE/CVF Conference on Computer Vision
  and Pattern Recognition}, pages 779--788, 2016.

\bibitem[Ren et~al.(2015)Ren, He, Girshick, and Sun]{ren2015faster}
Shaoqing Ren, Kaiming He, Ross Girshick, and Jian Sun.
\newblock {Faster R-CNN:} towards real-time object detection with region
  proposal networks.
\newblock In \emph{Proceedings of the Advances in Neural Information Processing
  Systems}, pages 91--99, 2015.

\bibitem[Saha et~al.(2017)Saha, Singh, and Cuzzolin]{saha2017amtnet}
Suman Saha, Gurkirt Singh, and Fabio Cuzzolin.
\newblock Amtnet: Action-micro-tube regression by end-to-end trainable deep
  architecture.
\newblock In \emph{Proceedings of the IEEE/CVF International Conference on
  Computer Vision}, pages 4414--4423, 2017.

\bibitem[Singh et~al.(2017)Singh, Saha, Sapienza, Torr, and
  Cuzzolin]{singh2017online}
Gurkirt Singh, Suman Saha, Michael Sapienza, Philip~HS Torr, and Fabio
  Cuzzolin.
\newblock Online real-time multiple spatiotemporal action localisation and
  prediction.
\newblock In \emph{Proceedings of the IEEE/CVF International Conference on
  Computer Vision}, pages 3637--3646, 2017.

\bibitem[Song et~al.(2019)Song, Zhang, Yu, and Sun]{song2019tacnet}
Lin Song, Shiwei Zhang, Gang Yu, and Hongbin Sun.
\newblock {TACNet:} transition-aware context network for spatio-temporal action
  detection.
\newblock In \emph{Proceedings of the IEEE/CVF Conference on Computer Vision
  and Pattern Recognition}, pages 11987--11995, 2019.

\bibitem[Soomro et~al.(2012)Soomro, Zamir, and Shah]{khurram2012ucf}
Khurram Soomro, Amir~Roshan Zamir, and Mubarak Shah.
\newblock {UCF101:} {A} dataset of 101 human actions classes from videos in the
  wild.
\newblock \emph{arXiv preprint arXiv: 1212.0402}, 2012.

\bibitem[Sun et~al.(2018)Sun, Shrivastava, Vondrick, Murphy, Sukthankar, and
  Schmid]{sun2018actor}
Chen Sun, Abhinav Shrivastava, Carl Vondrick, Kevin Murphy, Rahul Sukthankar,
  and Cordelia Schmid.
\newblock Actor-centric relation network.
\newblock In \emph{Proceedings of the European Conference on Computer Vision},
  pages 318--334, 2018.

\bibitem[Tang et~al.(2020)Tang, Xia, Mu, Pang, and Lu]{tang2020asynchronous}
Jiajun Tang, Jin Xia, Xinzhi Mu, Bo~Pang, and Cewu Lu.
\newblock Asynchronous interaction aggregation for action detection.
\newblock In \emph{Proceedings of the European Conference on Computer Vision},
  pages 71--87, 2020.

\bibitem[Tian et~al.(2019)Tian, Shen, Chen, and He]{tian2019fcos}
Zhi Tian, Chunhua Shen, Hao Chen, and Tong He.
\newblock Fcos: Fully convolutional one-stage object detection.
\newblock In \emph{Proceedings of the IEEE/CVF International Conference on
  Computer Vision}, pages 9627--9636, 2019.

\bibitem[Wang et~al.(2016)Wang, Qiao, Tang, and Van~Gool]{wang2016actionness}
Limin Wang, Yu~Qiao, Xiaoou Tang, and Luc Van~Gool.
\newblock Actionness estimation using hybrid fully convolutional networks.
\newblock In \emph{Proceedings of the IEEE/CVF Conference on Computer Vision
  and Pattern Recognition}, pages 2708--2717, 2016.

\bibitem[Weinzaepfel et~al.(2015)Weinzaepfel, Harchaoui, and
  Schmid]{weinzaepfel2015learning}
Philippe Weinzaepfel, Zaid Harchaoui, and Cordelia Schmid.
\newblock Learning to track for spatio-temporal action localization.
\newblock In \emph{Proceedings of the IEEE/CVF International Conference on
  Computer Vision}, pages 3164--3172, 2015.

\bibitem[Wu et~al.(2019)Wu, Feichtenhofer, Fan, He, Krahenbuhl, and
  Girshick]{wu2019long}
Chao-Yuan Wu, Christoph Feichtenhofer, Haoqi Fan, Kaiming He, Philipp
  Krahenbuhl, and Ross Girshick.
\newblock Long-term feature banks for detailed video understanding.
\newblock In \emph{Proceedings of the IEEE/CVF Conference on Computer Vision
  and Pattern Recognition}, pages 284--293, 2019.

\bibitem[Wu et~al.(2020)Wu, Kuang, Wang, Zhang, and Wu]{wu2020contextaware}
Jianchao Wu, Zhanghui Kuang, Limin Wang, Wayne Zhang, and Gangshan Wu.
\newblock {Context-Aware RCNN:} a baseline for action detection in videos.
\newblock In \emph{Proceedings of the European Conference on Computer Vision},
  pages 440--456, 2020.

\bibitem[Xie et~al.(2017)Xie, Girshick, Doll\'{a}r, Tu, and He]{Xie2016}
Saining Xie, Ross Girshick, Piotr Doll\'{a}r, Zhuowen Tu, and Kaiming He.
\newblock Aggregated residual transformations for deep neural networks.
\newblock In \emph{Proceedings of the IEEE/CVF Conference on Computer Vision
  and Pattern Recognition}, pages 5987--5995, 2017.

\bibitem[Yang et~al.(2019{\natexlab{a}})Yang, Yang, Liu, Xiao, Davis, and
  Kautz]{yang2019step}
Xitong Yang, Xiaodong Yang, Ming-Yu Liu, Fanyi Xiao, Larry~S Davis, and Jan
  Kautz.
\newblock Step: Spatio-temporal progressive learning for video action
  detection.
\newblock In \emph{Proceedings of the IEEE/CVF Conference on Computer Vision
  and Pattern Recognition}, pages 264--272, 2019{\natexlab{a}}.

\bibitem[Yang et~al.(2019{\natexlab{b}})Yang, Liu, Hu, Wang, and
  Lin]{RepPointsv1}
Ze~Yang, Shaohui Liu, Han Hu, Liwei Wang, and Stephen Lin.
\newblock {RepPoints}: Point set representation for object detection.
\newblock In \emph{Proceedings of the IEEE/CVF International Conference on
  Computer Vision}, pages 9657--9666, 2019{\natexlab{b}}.

\bibitem[Zhang et~al.(2019)Zhang, Tokmakov, Hebert, and Schmid]{zhang2019a}
Yubo Zhang, Pavel Tokmakov, Martial Hebert, and Cordelia Schmid.
\newblock A structured model for action detection.
\newblock In \emph{Proceedings of the IEEE/CVF Conference on Computer Vision
  and Pattern Recognition}, pages 9975--9984, 2019.

\bibitem[Zhao and Snoek(2019)]{zhao2019dance}
Jiaojiao Zhao and Cees~GM Snoek.
\newblock Dance with flow: Two-in-one stream action detection.
\newblock In \emph{Proceedings of the IEEE/CVF Conference on Computer Vision
  and Pattern Recognition}, pages 9935--9944, 2019.

\bibitem[Zhou et~al.(2019)Zhou, Wang, and Kr\"{a}henb\"{u}hl]{zhou2019objects}
Xingyi Zhou, Dequan Wang, and Philipp Kr\"{a}henb\"{u}hl.
\newblock Objects as points.
\newblock \emph{arXiv preprint arXiv: 1904.07850}, 2019.

\bibitem[Zhou et~al.(2020)Zhou, Koltun, and
  Kr\"{a}henb\"{u}hl]{zhou2020tracking}
Xingyi Zhou, Vladlen Koltun, and Philipp Kr\"{a}henb\"{u}hl.
\newblock Tracking objects as points.
\newblock In \emph{Proceedings of the European Conference on Computer Vision},
  pages 474--490, 2020.

\end{thebibliography}
\end{document}


\maketitle

In this supplementary material, we provide more experimental results, ablation studies, and visualizations to evaluate the performance of our Point3D in a comprehensive manner.

\section{Experimental Results}

\begin{figure}[!htb]
\setlength{\abovecaptionskip}{-1.5em}
\setlength{\belowcaptionskip}{-1em}
		\centerline{\includegraphics[width=0.7\linewidth]{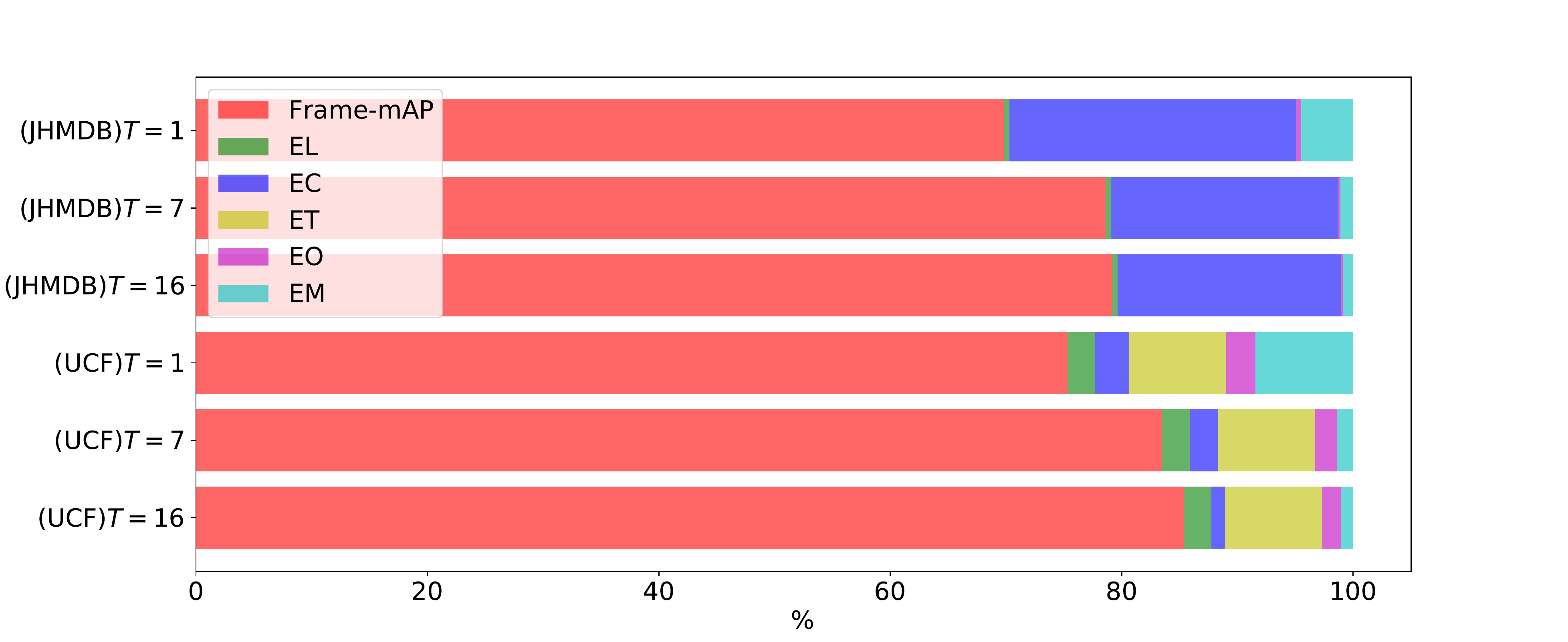}}
    	\caption{ Error analysis of our Point3D for $T = 1,7,16$
on the JHMDB and UCF101-24  dataset. We show frame-mAP  and different sources of error.}
\vspace{-0.5em}
	\label{fig: exp_error}
\end{figure}

In this section, we closely follow \cite{li2020actions,kalogeiton2017action} to conduct an error analysis on the frame mAP in order to better explore our proposed Point3D. Specifically, we investigate five kinds of action detection errors as described in \cite{kalogeiton2017action}, which are localization errors (EL), classification errors (EC), time errors (ET), other errors (EO), and missed detection errors (EM). 
Among these action detection errors, EL, EC, ET, and EO identify false positive detection and we follow the calculation of frame mAP and measure the area under the curve when plotting the percentage of each category at all recall values.
On the other hand, EM refers to the actions that we fail to detect at all. 
It is computed by measuring the percentage of ground truth boxes for which there are no correct detections.
We report the results for $T=1, 7, 16$ on the JHMDB and UCF$101$-$24$ dataset in Figure~\ref{fig: exp_error}. 

From Figure \ref{fig: exp_error}, we make the following observations: 
\textit{First}, when experimenting with $T=7$ on the JHMDB dataset, we achieve lower EL, EC, EO, and EM than the MOC~\cite{li2020actions} by $0.09\%$, $5.78\%$, $0.07\%$, and $0.20\%$, respectively. Our Point3D outperforms the MOC by $6.14\%$ in terms of frame-mAP. Similar improvements can be seen on the UCF$101$-$24$ dataset. This shows the advantage of our Point3D in spatio-temporal action recognition. 
\textit{Second}, with the increase of $T$, i.e., the length of the input clip, the frame-mAP increases and all the errors except ET decreases. This agrees with the common sense that the length of the input clip is crucial for action recognition. But there is a trade-off between the precision and the speed using different $T$s.
\textit{Third}, the increasing gap from $T=1$ to $T=7$ is insignificant compared to the increasing magnitude from $T=1$ and $T=7$. Thus, we set $T=7$ in our case for a better trade-off between the precision and the speed. From the error analysis, we can observe that our classification error EC (see the blue bar in Figure~\ref{fig: exp_error} is still high. 
Our Point3D sometimes makes some false positive classifications at initial frames since it is hard to recognize the action before it happens.



\begin{figure}[!htb]
\setlength{\abovecaptionskip}{-1.5em}
\setlength{\belowcaptionskip}{-1em}
		\centering
		\includegraphics[width=0.5\linewidth]{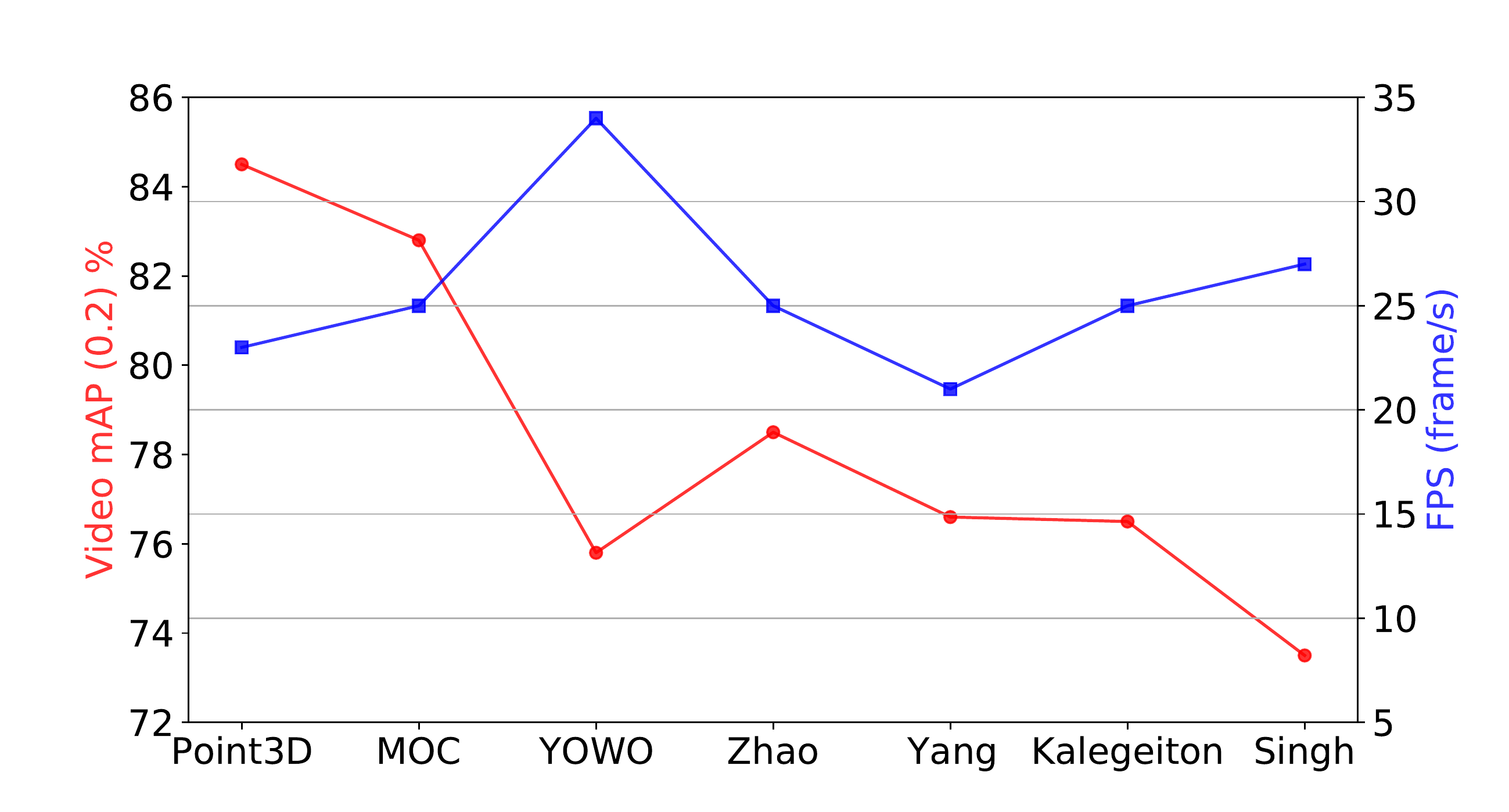}
    	\caption{Runtime comparison with the state-of-the-art methods using video mAP with a threshold of 0.2 and FPS of frame per second. A high video mAP (in red) and a small FPS (in blue) indicates a better performance. Results of \cite{saha2017amtnet} are omitted as their video mAP is much lower compared to other methods. }
	\label{fig: exp_runtime}
\end{figure}

Following previous work~\cite{li2020actions} closely, we also evaluate the two-stream offline speed of our Point3D's on a single Tesla V100 GPU. Point3D reaches a competitive speed of 20 fps compared with existing 2D detectors. In Figure~\ref{fig: exp_runtime}, we compare our Point3D with some current state-of-the-art methods which have reported their speed in the original paper~\cite{li2020actions, kopuklu2019yowo, zhao2019dance, yang2019step, kalogeiton2017action, singh2017online, saha2017amtnet}. Point3D achieves the best performance against the existing methods in terms of video mAP within reasonable runtime. This further confirms the advantages of our Point3D in tracking action as moving points with 3D-CNNs. 

\section{Ablation Study}

\noindent{\textbf{Input of 3D Head.}}
In this ablation study, we explore three types of 3D Head input, including the raw clip, frame heatmaps and the output from Point Head, as shown in Table \ref{tab: ab_input3D}. Output from the Point Head achieves better performance against the raw clip and the heatmap alone, which shows the advantage of the Point Head. Combining two of three input types can improve the performance further. Especially, using the raw clip and the output from Point Head achieves the best performance among all input combinations in terms of four video-mAP metrics. Feeding all three types of input into the 3D Head achieves the best frame-mAP, which further demonstrate the flexibility of the proposed Point3D architecture.

\begin{table}[h]
	\caption{Exploration study on the input of 3D Head.}
	\label{tab: ab_input3D}
	\renewcommand\tabcolsep{2.0pt}
	\centering
	\scalebox{0.8}{
		\begin{tabular}{c|c|c|c|c|c|c|c}
			\hline
			\multicolumn{1}{c|}{{Raw}} &  \multicolumn{1}{c|}{\multirow{2}{*}{Heatmap}}&
			\multicolumn{1}{c|}{{Point Head}}&
			\multicolumn{1}{c|}{Frame-mAP(\%)} & \multicolumn{4}{c}{Video-mAP(\%)}                     \\ \cline{5-8} 
\multicolumn{1}{c|}{clip}  & & \multicolumn{1}{c|}{output}                        & \multicolumn{1}{c|}{0.5}                                & 0.2  & 0.5  & 0.75 & \multicolumn{1}{c}{0.5:0.95} \\ \hline\hline
			\checkmark & &  & 76.1 & 87.4 & 84.5 & 69.6 & 58.7 \\

			& \checkmark & & 77.9 & 87.9 & 84.8 & 69.9 & 59.1 \\
			&   & \checkmark& 79.2 & 89.1 & 86.1 & 71.5 & 60.9 \\

			\checkmark & \checkmark & & 78.1 & 89.0   & 85.2 & 70.3 & 59.4 \\

			  & \checkmark & \checkmark &  79.4 & 89.3 & 86.2 & 71.7 & 61.1 \\
			 \checkmark&  & \checkmark & 79.5 & \textbf{89.4} & \textbf{86.5} & \textbf{71.8} & \textbf{61.3} \\
			 \checkmark & \checkmark & \checkmark &  \textbf{79.6} & 89.2 & 86.1 & 71.5 & 61.0 \\
			\hline
			\end{tabular}}
\end{table}

\noindent{\textbf{3D Head backbone.}}
Furthermore, we explore the design of 3D Head backbone, as shown in Table \ref{tab: ab_3Dbackbone}. Specifically, we employ the 3D backbone with 3D-ResNet with different depths \cite{targ2016resnet} and with other popular 3D-CNN architectures such as MobileNets \cite{howard2017mobilenets, sandler2018mobilenetv2} and ShuffleNet \cite{zhang2018shufflenet, ma2018shufflenet}. We can observe that the stronger 3D Head backbone we use, better the achieved results. This further demonstrates the importance of the proposed 3D Head in our Point3D. 
Moreover, applying the light-weight backbone in our 3D Head achieves the competitive performance compared to the MOC~\cite{li2020actions}. 

\begin{table}[h]
	\caption{Exploration study on the design of 3D Head backbone.}
	\label{tab: ab_3Dbackbone}
	\renewcommand\tabcolsep{4.0pt}
	\centering
	\scalebox{0.8}{
		\begin{tabular}{c|c|c|c|c|c|c}
			\hline
			\multicolumn{1}{c|}{\multirow{2}{*}{3D backbone}} &
			\multicolumn{1}{c|}{\multirow{2}{*}{FPS}} &\multicolumn{1}{c|}{Frame-mAP(\%)} & \multicolumn{4}{c}{Video-mAP(\%)}                     \\ \cline{4-7} 
\multicolumn{1}{c|}{}   & \multicolumn{1}{c|}{}                    & \multicolumn{1}{c|}{0.5}                                & 0.2  & 0.5  & 0.75 & \multicolumn{1}{c}{0.5:0.95} \\ \hline\hline
            MobileNetV1 2.0x  & \textbf{31}& 67.5 & 72.6 & 71.8 & 64.9 & 53.8 \\
			MobileNetV2 2.0x & \textbf{31}& 69.1 & 75.5 & 73.6 & 66.8 & 55.6 \\
			ShuffleNetV1 2.0x &30 &70.2 & 77.6 & 75.8 & 67.5 & 56.4 \\
			ShuffleNetV2 2.0x &30& 70.8 & 78.9 & 77.5 & 68.1 & 57.6 \\\hline
			ResNet-18 & 27 & 72.6 & 80.4 & 78.5 & 68.4 & 58.3 \\
			ResNet-50 & 25 & 74.5 & 82.5 & 80.3 & 68.7 & 58.8 \\
			ResNet-101 & 20 & 77.4 & 86.5 & 84.2 & 69.4 & 59.8 \\
			ResNeXt-101 & 23 & \textbf{79.2} & \textbf{89.1} & \textbf{86.1} & \textbf{71.5} & \textbf{60.9} \\
			\hline
			\end{tabular}}
\end{table}

\noindent{\textbf{Weight of localization and classification loss.}}
In Table \ref{tab: ab_weight}, we ablate $\lambda_{loc}$ and $\lambda_{cls}$, two important parameters in our Point3D to balance the weight of Point Head and 3D Head in the overall loss $L_{overall}$. We see that the performance gap among different settings are small, which further shows the robustness of our Point3D to the weight hyper-parameters. In our case, we set $\lambda_1=10$ and $\lambda_2=1$ to achieve the best performance. 

\begin{table}[h]
	\caption{Exploration study on $\lambda_{loc}$ and $\lambda_{cls}$.}
	\label{tab: ab_weight}
	\renewcommand\tabcolsep{6.0pt}
	\centering
	\scalebox{0.8}{
		\begin{tabular}{c|c|c|c|c|c|c}
			\hline
			\multicolumn{1}{c|}{\multirow{2}{*}{$\lambda_{loc}$}} & \multirow{2}{*}{$\lambda_{cls}$} & \multicolumn{1}{c|}{Frame-mAP(\%)} & \multicolumn{4}{c}{Video-mAP(\%)}                     \\ \cline{4-7} 
\multicolumn{1}{c|}{}                &       & \multicolumn{1}{c|}{0.5}                                & 0.2  & 0.5  & 0.75 & \multicolumn{1}{c}{0.5:0.95} \\ \hline\hline
            1 & 1  &79.2 & 89.1 & 86.1 & 71.5 & 60.9 \\
            1 & 5  &79.3 & 89.2 & 86.2 & 71.6 & 61.1 \\
            1 & 10 & 79.5 & 89.4 & 86.4 & 71.8 & 61.4 \\
			1 & 15  & 79.1 & 88.9 & 85.9 & 71.3 & 60.8 \\
			5  & 1 & 79.7 & 89.5 & 86.4 & 71.7 & 61.3 \\
			10 & 1 & \textbf{79.9} & \textbf{89.8} & \textbf{86.8} & \textbf{71.9} & \textbf{61.5} \\
			15 & 1 & 79.6 & 89.4 & 86.3 & 71.6 & 61.2 \\
			\hline
			\end{tabular}}
\end{table}

\section{Visualizations}

\begin{figure}[!htb]
		\centerline{\includegraphics[width=0.67\linewidth]{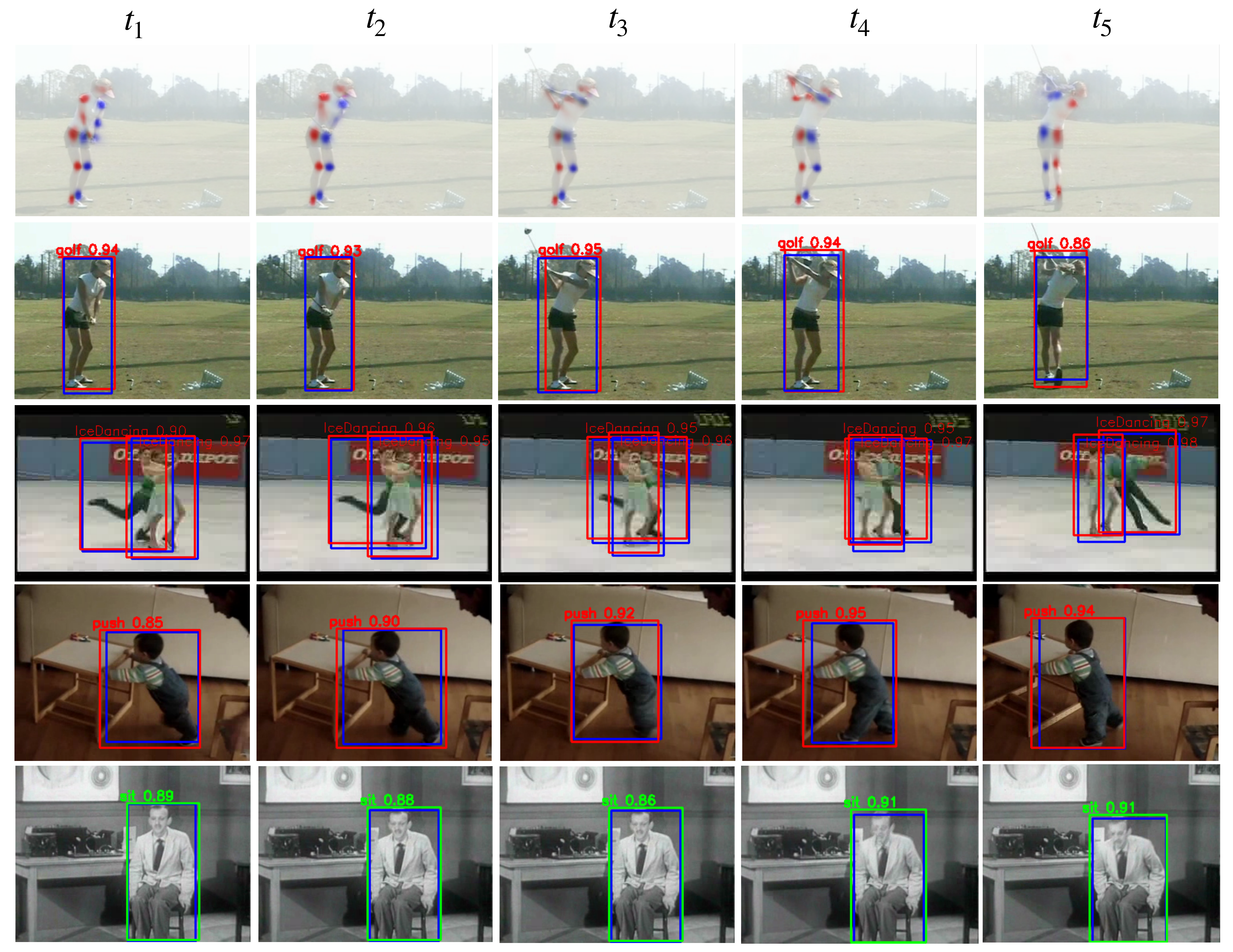}}
    	\caption{\textbf{Visualization results of how our Point3D conducts action recognition and some examples from JHMDB and UCF101-24.} The first row denotes the heatmaps generated from our KP detector, tracking spatial and temporal changes of the action from frame to frame. The second, third, and fourth rows show examples of true positive detections. The last row shows an false positive detection example that the action ``stand up" is misclassified into ``sit" but the localization of this action is still robust. The \textbf{blue} bounding boxes are ground truths while the \textbf{red} and \textbf{green} boxes are true and false positive detections, respectively. Zoom in for a better view.}
	\label{fig: exp_vis_all}
\end{figure}

\begin{figure}[!htb]
		\centerline{\includegraphics[width=0.9\linewidth]{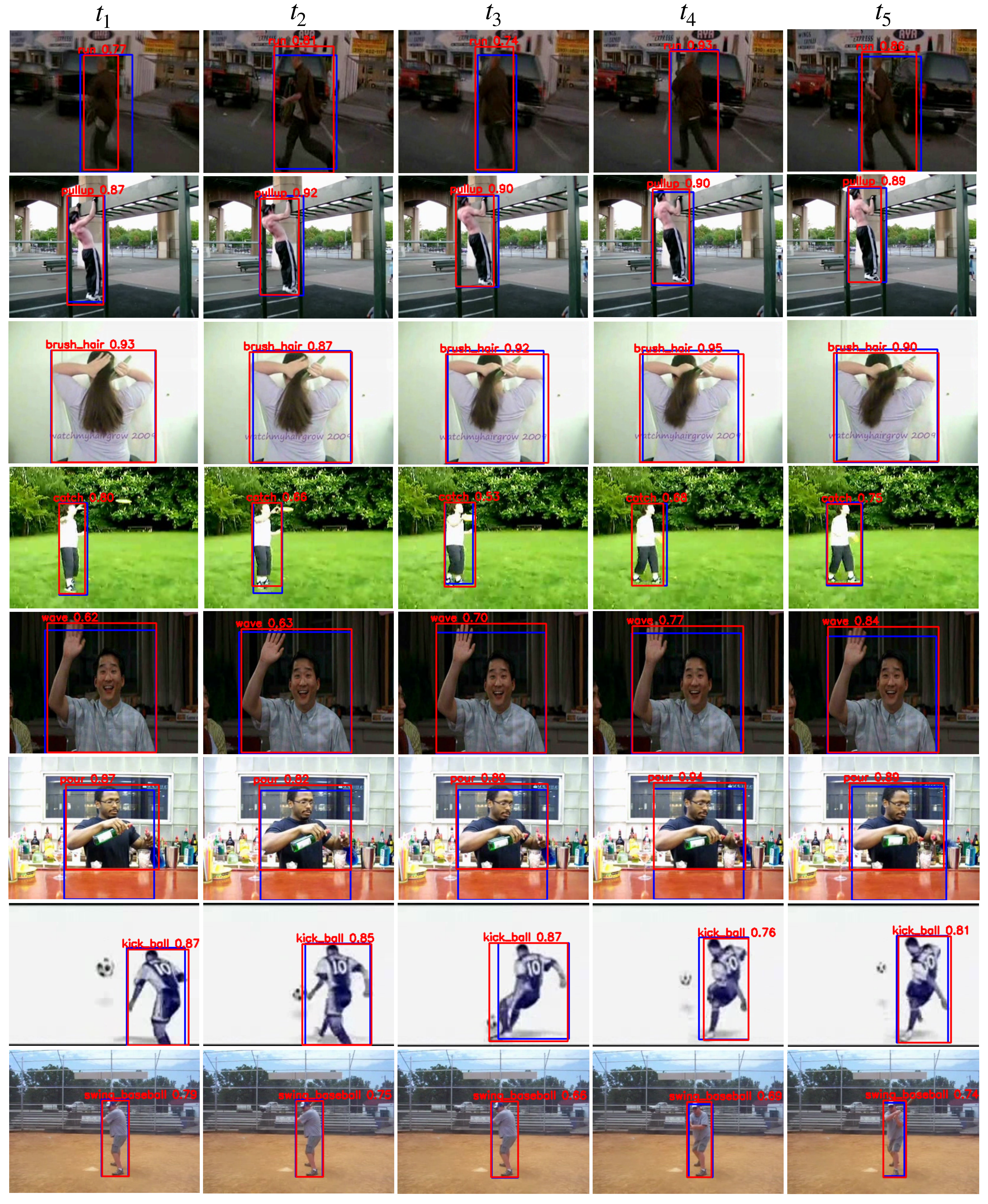}}
    	\caption{\textbf{Visualization results of examples from JHMDB}. The \textbf{blue} bounding boxes are ground truths while the \textbf{red} are true positive detections, respectively. The actions from top to bottom are ``run'', ``pullup'', ``brush hair'', ``catch'', ``wave'', ``pour'', ``kick ball'', and ``swing baseball''. Zoom in for a better view.}
	\label{fig: exp_vis_jhmdb}
\end{figure}

\begin{figure}[!htb]
		\centerline{\includegraphics[width=0.9\linewidth]{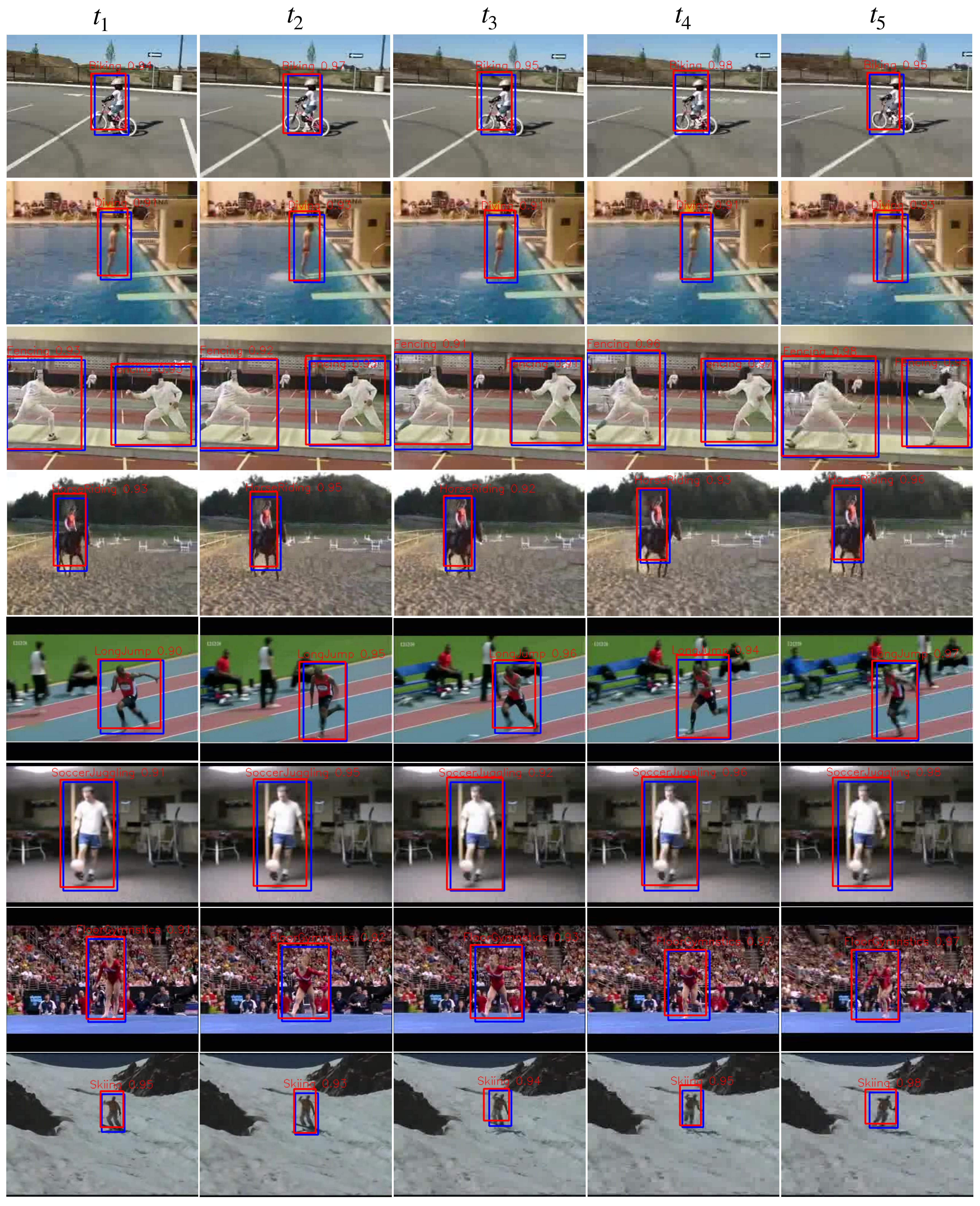}}
    	\caption{\textbf{Visualization results of examples from UCF101-24}. The \textbf{blue} bounding boxes are ground truths while the \textbf{red} are true positive detections, respectively. The actions from top to bottom are ``Biking'', ``Diving'', ``Fencing'', ``Horse Riding'', ``Long Jump'', ``Soccer Juggling'', ``Floor Gymnastics'', and ``Skiing''. Zoom in for a better view.}
	\label{fig: exp_vis_ucf}
\end{figure}

We provide more qualitative examples of action recognition on the JHMDB and UCF101-24 datasets to demenstrate the effectiveness of our proposed Point3D. In general, our Point3D architecture exhibits satisfactory performance at localizing and classifying actions in videos. As can be seen in the first row of Figure \ref{fig: exp_vis_all}, the heatmaps generated from our Knot-Point (KP) detector track spatial and temporal changes of the action from frame to frame, without needing any anchor boxes. 
For true positive examples, the actions are correctly classified and the bounding boxes predicted from Point3D are nearly identical to the ground truth bounding boxes. For false positive examples, our results show even though the action is misclassified, the localization result is still robust. We provide more visualization examples using videos in JHMDB and UCF101-24 in Figure \ref{fig: exp_vis_jhmdb} and Figure \ref{fig: exp_vis_ucf}, respectively.

\section{Codes to replicate the experiments}

We attach the codes to replicate our experiments and the usage of Point3D as part of supplementary material.

\bibliography{reference}


\title{Point3D: tracking actions as moving points with 3D CNNs \\ \textit{(Supplementary Material)}}


\author{First Author\\
Institution1\\
Institution1 address\\
{\tt\small firstauthor@i1.org}
\and
Second Author\\
Institution2\\
First line of institution2 address\\
{\tt\small secondauthor@i2.org}
}
\maketitle
\ificcvfinal\thispagestyle{empty}\fi






